\documentclass{article}
\usepackage{amsmath, amssymb, amsfonts, amsthm}
\usepackage{mathtools}
\usepackage{graphicx}
\usepackage{enumitem}
\usepackage{caption,setspace}
\usepackage{subcaption}
\usepackage{multirow}
\usepackage{wrapfig}

\usepackage{mathrsfs}
\usepackage{mathtools}
\usepackage[dvipsnames]{xcolor}
\usepackage{hyperref, url}
\usepackage{lipsum}
\usepackage{tikz-cd}

\usepackage[numbers,sort&compress,square]{natbib} 

\usepackage{enumitem} 

  {\begin{enumerate}%
    leftmargin = *
    \setlength{\itemsep}{0pt}%
    \setlength{\parskip}{0pt}}%
  {\end{enumerate}}

\theoremstyle{plain}
\newtheorem{theorem}{Theorem}[section]
\newtheorem{lemma}[theorem]{Lemma}

\newtheorem{proposition}[theorem]{Proposition}        
\theoremstyle{definition}

\theoremstyle{definition}
\newtheorem{Definition}[theorem]{Definition}
      
\theoremstyle{remark}

\usepackage[accepted]{icml2019}

\begin{document}

\twocolumn[
\icmltitle{Active Manifolds: A non-linear analogue to Active Subspaces}



\icmlsetsymbol{equal}{*}

\begin{icmlauthorlist}
\icmlauthor{Robert A. Bridges}{ornl}
\icmlauthor{Anthony D.  Gruber}{ttu,ornl}
\icmlauthor{Christopher R. Felder}{wustl,ornl}
\icmlauthor{Miki E. Verma}{ornl}
\icmlauthor{Chelsey Hoff}{ornl}
\end{icmlauthorlist}

\icmlaffiliation{ornl}{Cyber \& Applied Data Analytics Division, Oak Ridge National Laboratory}
\icmlaffiliation{ttu}{Department of Mathematics, Texas Tech University}
\icmlaffiliation{wustl}{Department of Mathematics and Statistics, Washington University in St. Louis}

\icmlcorrespondingauthor{Robert A. Bridges}{bridgesra@ornl.gov}

\icmlkeywords{approximation,  regression, dimension reduction, sensitivity analysis, parameter studies}

\vskip 0.3in
]



\printAffiliationsAndNotice{
\scriptsize{This manuscript has been co-authored by UT-Battelle, LLC, under contract DE-AC05-00OR22725 with the US Department of Energy (DOE). The US government retains and the publisher, by accepting the article for publication, acknowledges that the US government retains a nonexclusive, paid-up, irrevocable, worldwide license to publish or reproduce the published form of this manuscript, or allow others to do so, for US government purposes. DOE will provide public access to these results of federally sponsored research in accordance with the DOE Public Access Plan (http://energy.gov/downloads/doe-public-access-plan).\\
}
}  


\begin{abstract} \small\baselineskip=9pt 
We present an approach to analyze  $C^1(\mathbb{R}^m)$ functions 
that addresses limitations present in the Active Subspaces (AS) method of Constantine et al. \citeyearpar{constantine2013active, constantine2015active}. 
Under appropriate hypotheses, our Active Manifolds (AM) method  identifies a 1-D curve in the domain (the active manifold) on which nearly all values of the unknown function are attained, and which can be exploited for approximation or analysis, especially when $m$ is large  (high-dimensional input space).  
We provide theorems justifying our AM technique and an algorithm permitting functional approximation and sensitivity analysis. 
Using accessible, low-dimensional functions as initial examples, we show AM reduces approximation error by an order of magnitude compared to AS, 
at the expense of more computation.
Following this, we revisit the sensitivity analysis by Glaws et al. \citeyearpar{glaws2017dimension}, 
who apply AS to analyze a magnetohydrodynamic power generator model, 
and compare the performance of AM on the same data.  
Our analysis provides detailed information not captured by AS, exhibiting
the influence of each parameter individually along an active manifold.
Overall, 
AM represents a novel technique for analyzing functional models with benefits including: reducing 
$m$-dimensional analysis to a 1-D analogue,
 permitting more accurate regression than AS 
(at more computational expense), 
enabling more informative sensitivity analysis, and granting accessible visualizations (2-D plots) of parameter sensitivity along the AM. 
\end{abstract}

\section{Introduction}
Scientists and engineers rely on accurate mathematical models to quantify the objects of their studies.
Such models can be difficult to analyze because they appear as systems of implicitly defined equations, or  they involve a high number of parameters relative to the quantity of data/observations.
For example, regression to determine a high number of parameters requires a large amount of data, else the problem is under-determined.
Similarly, models are frequently 
subject to questions of sensitivity, and to prioritize next step research it is desirable to answer questions such as  ``According to the mathematical model, which parameters should we change to get the most response in the output?'', e.g., \cite{constantine2013active, constantine2015exploiting,
lukaczyk2014active, constantine2016accelerating, constantine2017global}.

To this end, we consider the problem of analyzing a real-valued function that is  $C^1$ (at least one continuous derivative) defined on a problematically high-dimensional domain, with the aim of performing dimension reduction and sensitivity analysis. 
One way to address this problem is to tailor methods of dimension reduction to the given model, in order to work in a smaller parameter space while preserving fidelity of the output. 
Essentially, this entails approximating the original model using fewer inputs. 

Given a $C^1$ function, $f:U \subset \mathbb{R}^m \to \mathbb{R}$ with potentially large $m$, we fashion our approach on a simple concept from vector calculus.
Suppose one is standing at a point $x_0\in U$, observes the value $c = f(x_0)$, and takes a small step.
Then, there are $m-1$ orthogonal directions (potentially a huge subspace!) that one can step without changing the function (or more precisely, while exhibiting negligible change in $f$).
Yet, there is one special direction, the gradient $\nabla f_{x_0}$, and so long as the step is in this direction, the walker is guaranteed maximal change in $f$ near $x_0$!
More formally, we are simply stating that the $m$-dimensional tangent space to $\mathbb{R}^m$ at $x_0$ can be decomposed into an $(m-1)$-dimensional subspace of vectors tangent to the level set $\{x: f(x) = c\}$, and a one-dimensional space consisting of scalar multiples of the gradient $\nabla f_{x_0}$.
Armed with this perspective, we explore a powerful hypothesis\textemdash by exploiting this decomposition, we can reduce analysis of any $f \in C^1(\mathbb{R}^m)$ to a related $\hat f \in C^1(\mathbb{R})$, thereby allowing for analysis of arbitrarily high-dimensional models in a single dimension.

\begin{wrapfigure}[15]{r}{.24\textwidth}
\hspace{-0.8pc}
\centering
    \includegraphics[scale = .115]{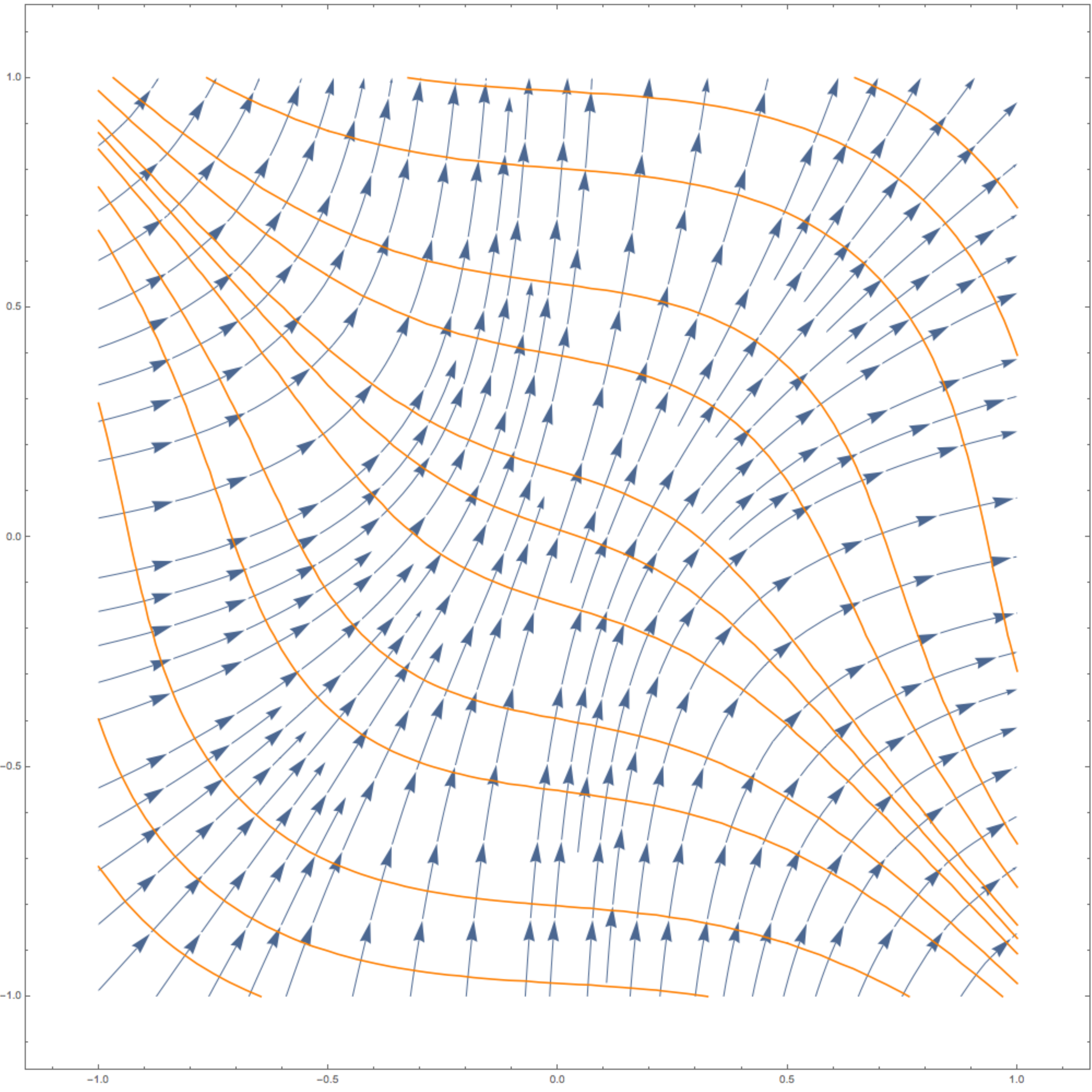}
    \vspace{-0.6pc}
    \caption{\textcolor{orange}{Level sets (orange)} of example function $f_3: \mathbb{R}^2 \to \mathbb{R}$, and its \textcolor{blue}{gradient vector field (blue)} tangent to the AMs at each point.}
    \label{fig:vectorfield}
\end{wrapfigure}

{\it Contributions: }
We provide the mathematical foundation and pseudo-algorithm for a novel method of analyzing functions with a problematically high-dimensional domain.
The method first recovers a curve in the domain, called the active manifold, and then reduces the problem to analysis of the function restricted to only this 1-D manifold by traversing level sets that necessarily intersect the active manifold orthogonally. E.g., see Figure \ref{fig:vectorfield}. 

This work is related to sliced inverse regression \cite{li1991sliced, duan1991slicing}, but is directly inspired by and builds upon the the Active Subspaces (AS) method of Constantine \citeyearpar{constantine2013active, constantine2015active},
and we adopt the name Active Manifolds (AM) in recognition.
Constantine's AS method seeks an affine subspace inside the domain on which the function changes most on average.
Unfortunately, AS does not guarantee there is a lower dimensional subspace to be produced by the algorithm (e.g., all $m$ directions can contribute equally to the response as in $f(x) = |x|^2$). 
Furthermore, because AS is a summary analysis over the whole domain, the potential for losing important information about the function is great; specifically, the impact of particular parameters on a model can be ``averaged out'' by AS.
Instead, we craft a non-linear analogue by considering iterative local analysis rather than global summary statistics.
We provide a rigorous mathematical foundation informing 
a psuedo-algorithm for function approximation using AM.

In accordance with AS, we assume we can sample both the function value $f(x)$ and the gradient $\nabla f_x := \nabla f(x)$ (or are given a random sample), and consider the problems of regression and sensitivity analysis.
We present experiments on known functions to illustrate the method in an easily understandable setting, producing regression results testing AM against AS. 
Further, we consider a real-world model of a magnetohydrodynamic (MHD) power generator. 
Following Glaws et al. \citeyearpar{glaws2017dimension} who used AS for sensitivity analysis of the MHD model,  
we now apply AM to see what additional information we can extract. 
Our results show that, even with their selected model and data, AM offers distinct advantages over AS at the expense of more computation.
In particular, we are able to improve estimation error as well as give a more detailed and interpretable characterization of the model's parameter sensitivity.


Benefits of the AM approach are fourfold:
(1) Regardless of the input dimension $m$, AM reduces the problem to analysis of a one-dimensional (1-D) function. 
In particular, this relieves the burden of AS on the user to choose a suitable subspace for analysis.
(2) In initial experiments we exhibit more accurate regression than the AS method with the same input data, though at greater computational expense.
(3) Our method allows for finer and more informative sensitivity analysis by providing local rather than global rankings of the input parameters based on their influence. 
This permits segmenting the active manifold 
into regions where sensitivity is different.
AM also allows the user to see the influence of each parameter  \emph{individually} along the manifold, contrasting with AS which gives sensitivity rankings globally in terms of a (perhaps physically meaningless) linear combination of parameters.
(4) The 1-D approach of AM allows accessible visualization (2-D plots) to inform understanding of the high-dimensional function. 


\section{Related Works}
\label{sec:related}

Dimension reduction, broadly speaking, is the mapping of data to a lower dimensional space, with the goal of preserving and illuminating some desired characteristics by eliminating unneeded degrees of freedom.
See Burges \citeyearpar{burges2010dimension} for an overview. 
Powerful and well-known techniques include Principal Component Analysis  \cite{PCA}, the Nystr\"{o}m method \cite{kumar2012sampling}, Isomap \cite{zhang2013misomap}, Diffusion Maps \cite{de2008introduction}, and Norm Discriminant Manifold Learning \cite{liu2018l_}. 

Sliced Inverse Regression (SIG)  \cite{li1991sliced, duan1991slicing} and later refinements, e.g., \cite{li06sparce, coudret142new}, are related to our AM method. 
The main idea is to model $y:\mathbb{R}^m\to\mathbb{R}$ as $y = f(Bx + \epsilon)$ for unknown $f$ and Gaussian noise $\epsilon$.
The goal is to learn $B$, an $m \times m$ matrix with rank $k\leq m$, which gives a lower dimensional subspace on which $y$ is recoverable. 
Roughly speaking, Li provides a theorem that states  $x\sim N(0,I)$ yields $\mathbb{E}(x|y)$ is in range$(B)$, which informs an algorithm: normalize $x$;  empirically estimate $\mathbb{E}(x|y)$; use the SVD to recover $B$ from the $k$ most significant directions.

Closely related to SIG and a primary driver for this work is Active Subspaces (AS), an idea originally of Russi \citeyearpar{russi2010uncertainty} but developed extensively by  Constantine et al. \citeyearpar{constantine2013active, constantine2015active}. See Sec. \ref{sec:as}. 
Many applications have been found for AS, e.g., shape optimization \cite{lukaczyk2014active}, MCMC for Bayesian inverse problems \cite{constantine2016accelerating}, and sensitivity analysis \cite{constantine2017global}. 

Emerging research of Zhang \& Hinkle \citeyearpar{zhang2019resnet} builds on the AM idea of this paper by using ResNets to learn a (generally) non-linear, lower-dimensional transformation of the input variable $x$ with minimal reconstruction error of both the desired function and gradient.  
This can be considered an analogue to AM with tunable dimension, by using the ResNet statistical machinery to learn the manifold.

A common challenge of the above methods is deciding the dimension of the reduced space. 
It is often necessary to inspect eigenvalues and make an educated guess about what dimension is needed to capture important information.  
AM avoids this issue by projecting the relevant features to 1-D in every case.
Although the work presented here departs from the use of traditional projective and spectral methods, it is still a manifold modeling method.

\subsection{Active Subspaces (AS)}
\label{sec:as} 
Developed by Constantine et al.~\citeyearpar{constantine2013active}, AS is a dimension reduction technique that is both applicable to a wide class of functions (those with $C^1$ regularity) and accessible to scientists and engineers with limited mathematical background.
Even better, AS is fast, because it focuses on affine approximation where linearity can be exploited.
In particular, AS looks for a lower-dimensional affine subspace inside the domain of a function $f$ by computing the directions in which $f$ changes the most on average.
As nice as this is, the AS method has three main limitations: first, the inherent restriction present in considering only affine subspaces can create large errors, as we will see in later examples;
secondly, visualization can be an issue with this method, since the active subspaces themselves are not guaranteed to be low-dimensional;
finally, many functions do not admit an active subspace, e.g. $f(x) = |x|^2$ as mentioned earlier.

Below is a heuristic description of the AS algorithm.
See \cite{constantine2013active, constantine2015active} for more details.
Assume $f\in C^1([-1,1]^m)$.
\vspace{-0.5pc}
\begin{enumerate}[leftmargin = *,topsep=5pt,itemsep=2pt,parsep=4pt]
\item Sample $\nabla f_{a_i}:= \nabla f(a_i)$ and $f(a_i)$ at $N$ random points $a_i \in [-1,1]^m$.
\item 
Compute the eigenvalue decomposition of the matrix
\vspace{-.2cm}
$$\mathbf{C} = \frac{1}{N} \sum_{i=1}^{N} \nabla f_{a_i} \nabla f_{a_i}^{T} = \mathbf{W}\mathbf{\Lambda}\mathbf{W}^{T}.$$
\vspace{-.5cm}
\item Manually inspect the set of eigenvalues $\{\lambda_i\}$ for ``large" gaps. If there is a gap between $\lambda_j$ and $\lambda_{j+1}$, we say $S := \text{Span}\big(\{w_1,...,w_j\}\big)$ is the ``active subspace'' of dimension $j$ associated to $f$, where the $w_i$ are the corresponding eigenvectors.
\item Given an arbitrary point $p \in [-1,1]^m$, project $p$ orthogonally to $p': = Proj_{S}(p)$ on the active subspace.
\item  Define $\hat f: S\to \mathbb{R}$ by $\hat f(x) := \text{ave}\{f(a_i): Proj_{S}(a_i)$ near $x_0\}$ and obtain the approximation $f(p) \approx \hat f(p')$. 
\end{enumerate}
The AS algorithm also allows for a bootstrapping procedure, in which active subspaces are computed from random partitions of the input data.
Comparing the range of these to the full active subspace gives ``confidence'' values that are usually plotted along with the values of $f$ on the active subspace. 
See Constantine~\citeyearpar{constantine2015active} for more details.  

\section{Active Manifolds (AM)}
\label{sec:AM}
Here we provide the mathematical foundation for AM and describe a pseudo-algorithm for reducing analysis of the $m$-dimensional function to its one-dimensional analogue.
Examples to illustrate the method are provided, including illustrations of problems or obstructions identified.
\vspace{-0.4pc}
\subsection{Mathematical Justification: }
Recall that the arc length of a  $C^1$ curve $\gamma(t) : [0, 1] \rightarrow \mathbb{R}^m$ is given by
 $S(\gamma) = \int_0^1 |\gamma'(t)| \ dt.$
Let $U \subset \mathbb{R}^m$ open and assume $f\in C^1(U)$.

We seek
\vspace{-.1cm}
$$\arg \max \int_{0}^{1}  \big\langle \nabla f(\gamma(t)), \gamma'(t) \big\rangle \ dt $$
over all $C^1$ curves $\gamma(t) : [0, 1] \rightarrow U$, such that $|\gamma'| = 1$ (constant speed), where $\langle\cdot,\cdot\rangle$ denotes the usual Euclidean inner product.
Note that the integrand satisfies
$$ \langle \nabla f(\gamma(t)), \gamma'(t) \rangle  = |\nabla f(\gamma(t)) | \ |\gamma'(t) | \cos{\theta} $$
where $\theta$ is the angle between $\nabla f(\gamma(t))$ and $\gamma'(t)$. Clearly this quantity is maximal when $\theta = 0$, indicating that $\nabla f(\gamma(t))$ and $\gamma'(t)$  point in the same direction; hence, the solution to this optimization problem is
\begin{equation}
\label{am-condition}
\gamma'(t)= \frac{\nabla f(\gamma(t))}{| \nabla f(\gamma(t)) |}\ ,
\end{equation}
a constant-speed streamline of $\nabla f$.
Specifying a starting point, $\gamma_0,$ uniquely identifies the flow as furnished by the following standard theorem of  differential equations.
\begin{lemma}
\label{am-prop-1}
Given $f : U \subset \mathbb{R}^m \xrightarrow[]{C^1} \mathbb{R}$ and an initial value $\gamma_{0} \in U$, there exists a unique local solution $\gamma(t)$ to the system of first-order ordinary differential equations described by (\ref{am-condition}).
\end{lemma}
\vspace{-1pc}
\begin{proof}
Choose any compact and convex subset $K \subset U$ containing $\gamma_0$. 
Since $f$ is $C^1$, $\nabla f$ satisfies the Lipschitz condition,
$| \nabla f(x_1) - \nabla f(x_2) | \leq L_K |x_1 - x_2|,$ for $x_1, x_2 \in K$\ and $L_K <\infty$ is some Lipschitz constant. 
By Theorem 1 Ch. 6 from Birkhoff \& Rota \citeyearpar{birkhoff1969ODE}, these conditions are sufficient for the existence and uniqueness of a local solution $\gamma(t)$ to Eqn.  (\ref{am-condition}) about $\gamma_0$ in $K$, which can be reparametrized to have domain $[0,1]$ as desired. Since $K$ was an arbitrary compact set we have the result. 
\end{proof}

\begin{Definition}{
Let $f : U \subset \mathbb{R}^m \xrightarrow[]{C^1} \mathbb{R}$. We say that $\mathcal{M} \subset U$ is an \textbf{active manifold} defined by $f$ provided there exists a constant-speed parametrization of $\mathcal{M}$, $\gamma(t):[0,1] \to U$, such that condition (\ref{am-condition}) is satisfied for all $t \in [0,1]$.}
\end{Definition}\label{am-def}

\vspace{-0.25pc}
\noindent For the following proposition, let $f, U$ be as above and:  \\
\\
\noindent\begin{minipage}{.28\textwidth}
\begin{itemize}[leftmargin = *,labelindent=0mm,labelsep=1mm, itemsep=1mm]
\item $\mathcal{M}$ = Im$\,\gamma(t)$ an active manifold of $f$
\item The relation $\sim$ defined by $f$, i.e., $\forall x,y \in U, \\ x\sim y \iff f(x) = f(y)$
\item $[x] = \{ y \in \mathbb{R}^m : f(x) = f(y) \}$
\item $\mathbb{R}^m/{\sim} = \{[x]: x\in \mathbb{R}^m\}$
\item $\pi : \mathbb{R}^m \twoheadrightarrow \mathbb{R}^m/{\sim}$
\end{itemize}
\end{minipage}
\hspace{-1.5pc}
\begin{minipage}{.18\textwidth}
\begin{tikzcd}[row sep=scriptsize, column sep = scriptsize]
 	& U \arrow[rd, "f"] \arrow[d, "\pi"] & \\[1.9cm]
	[0,1] \arrow[ur, "\gamma"] \arrow[r, "\pi \circ \gamma"] & U/{\sim} \arrow[r, "\hat{f}"] & \mathbb{R}
\end{tikzcd}
\end{minipage}



\vspace{0.5pc}

\begin{proposition}
\label{FToAM}
If $\gamma(t)$ is a solution to Eqn. (\ref{am-condition}) on an open set $U$ away from points where $\nabla f = 0$, then the following statements hold.
\vspace{-0.5pc}
\begin{enumerate}[label=(\roman*), leftmargin = *,topsep=5pt,itemsep=2pt,parsep=5pt]
\item $\mathcal{M}$ is an immersed $C^1$ submanifold of $U\subseteq \mathbb{R}^m$.
\item $U/\sim$ is a $C^1$ manifold.
\item $\pi\circ\gamma$ is a $C^1$ embedding of $\mathcal{M}$ in $\mathbb{R}^m/{\sim}$.
\item $f\circ\gamma: [0,1]\to \mathbb{R}$ is strictly  increasing. 
\end{enumerate}
\end{proposition}

\vspace{-1pc}
\begin{proof}
(i):  Note that $f \rvert_{\mathcal{M}}$ provides a global $C^1$ chart for $\mathcal{M}$. Further, $\mathcal{M}$ is immersed since $|\gamma'|=1$, hence $\gamma'$ does not vanish.
(ii):  Since $f$ is $C^1$ and constant on the fibers of $\pi$, the map $\hat f: U/{\sim} \to \mathbb{R}$ defined as $\hat{f}([x]) := f(x)$ is $C^1$. So $U/{\sim}$ is a $C^1$ manifold with global chart $\hat f$.
(iii):  $\pi\rvert_\mathcal{M}$ is a bijection onto $\pi(\mathcal{M})$ since $\mathcal{M}$ fibers pointwise under $\pi$. 
Since $\pi$ is linear, it follows that $d\pi\rvert_{\mathcal{M}} = \pi\rvert_{\mathcal{M}}$ is bijective; 
hence, $\pi$ is an embedding.
(iv): Monotonicity of $f\circ\gamma$ follows directly from the definition: $\forall t$ $\gamma(t) \parallel \nabla f(t)$. \end{proof}

\begin{theorem}
Suppose the level set $\{ f = \alpha \}$ is connected and $\gamma$ is any active manifold such that $\alpha \in$ Im$(f\circ\gamma)$. 
Then $\exists\, !\,\, t_0$ such that $\gamma \cap \{ f = \alpha \} = \{\gamma(t_0)\}$, and $\gamma \perp \{ f = \alpha \}$.
\end{theorem}
\vspace{-1pc}
\begin{proof}
The Implicit Function Theorem guarantees that for each $\alpha \in  \text{Im}\,f$, the level set $\{x: f(x) = \alpha \}$ is an $(m-1)$-dimensional submanifold of $\mathbb{R}^m$ that is orthogonal to the gradient vector field and therefore to any intersecting active manifold.  
By hypothesis $\exists \, t_0$ such that $\gamma(t_0)\in \{f=\alpha\}$. 
Uniqueness follows from monotonicity of $f\circ\gamma$ (Proposition \ref{FToAM}.\textit{iv}).
\end{proof}

\vspace{-0.5pc}
\textit{Implication: } This theorem implies that if one can recover $f\circ \gamma$  (a 1-D regression problem), then one can recover $f$ on the connected component of any level set touching $\gamma$. 
Concisely, if $p$ is in the component of $A := \{f=f(p)\}$ intersecting $\gamma$, one may move freely in the $(m-1)$-dimensional submanifold $A$ transverse to $\gamma$ without changing $f$. This motivates our AM pseudo-algorithm.


\vspace{-0.4pc}
\subsection{Active Manifolds Pseudo-Algorithm: }
The AM algorithm has three broad components:
(1) Build the active manifold $\mathcal{M} = $ Im$(\gamma(t))$; (2) Approximate the function of interest $f$ on $\mathcal{
M}$  with $\hat f$; (3) For $p\in {U}$ traverse the level set to $\mathcal{M}$  to estimate $f(p)$.
We require two parameters: $\delta$, a step size for the numerical approximation of paths, and $\epsilon$, a tolerance for when to terminate walking. 
We now discuss each component in detail.
\vspace{-0.25pc}
\subsubsection{Building the Active Manifold: }
\label{sec:build-am}
For the algorithm, we consider $f$ on the hypercube 
$U = [-1,1]^m$, assuming one has pre-composed $f$ with a scaling function on a portion of the original domain if necessary. 
Given starting value $p \in U$, we describe the process of building the active manifold, $\gamma$, 
which will be a one-dimensional curve in  $[-1,1]^m$ that moves from a local minimum of $f$ to a local maximum. 
As are the assumptions of AS, we require observational data of $f$ and $\nabla f$ at some samples in the domain. 
We first build a uniform grid with spacing size $\epsilon$ and compute $f, \nabla f$ at each grid point. 
We then use a gradient ascent/descent scheme with
 nearest neighbor search 
to construct the active manifold. 
We set the step size $\delta = 2d/3$ where $d$ is the longest diagonal of the hypercubes in our sampled grid. 
\textit{Note that a grid is not necessary, as one could easily rework the algorithm to accommodate other initial data, e.g., a given set of observations in the hypercube. 
See Algorithm Note~\ref{sec:notes}.\ref{itm:notes-2}.  }
\vspace{-0.5pc}
\begin{enumerate}[leftmargin = *, topsep=5pt,itemsep=2pt,parsep=5pt]
\item Build a uniform 
grid $\mathbf{P} = \{p_i\}$ over $[-1,1]^m$ with spacing size $\epsilon.$ Sample $f$ and $\nabla f$ at each grid point. 
See Algorithm Note \ref{sec:notes}.\ref{itm:notes-1}. 

\item  Given an initial starting point $\gamma_0 \in U$, find the nearest $p_i$, and set $\gamma_1:= \gamma_0 + \delta\, \nabla f(p_i)/|\nabla f(p_i)|$. 
Continue this gradient ascent/descent scheme in both directions to find a numerical solution to 
$\gamma'(t) = {\nabla f(\gamma(t))} / {|\nabla f(\gamma(t))|}$
using the samples $\{p_i, f(p_i), \nabla f_{p_i}\}$ from step 1. 
The algorithm in each direction ends when either the next step would exit $[-1,1]^m$ or would become close to a previous step (we use  $\delta/3$ as the closeness parameter). 
The set $\mathcal{M} = \{\gamma_i\}_i$ is then a discretized active manifold in $[-1,1]^m$. 
See Algorithm Note \ref{sec:notes}.\ref{itm:notes-3}. 

\item Finally we parameterize $\gamma$ on $[0,1]$ as follows: 
While the active manifold is built, save the number of steps $0\leq i \leq N$ and the function value $z_i = f(\gamma_i)$ at each step. Use this to construct ordered lists $\mathbf{S}:= \{i = 0, \dots, N\}$ and $\mathbf{Z}:= \{z_i\}$.
Scale $\mathbf{S}$ so that $\mathbf{S}: = \{i/N\}$ and the domain of $\gamma$ is $[0,1]$.  Note that such a parameterization is necessarily constant-speed.
\end{enumerate}

\vspace{-0.5pc}
\subsubsection{Approximating $f$ with $\hat f$: }
To obtain a one-dimensional approximation $\hat f \approx f$ defined on the whole of $[0,1]$, we fit the data $\{(i/N, z_i) \in \mathbf{S} \times \mathbf{Z} \}$ to a piecewise-cubic Hermite interpolating polynomial $\hat{f}: [0,1] \to \mathbb{R}$.  
Note that $\hat f$ is globally $C^1$ and monotone. 
This furnishes a benefit of our method\textemdash $\hat f$, which represents $f$ along the active manifold, can be plotted for a useful 2-D visualization of the data, e.g., Figures \ref{fig:f3amas} (left) \& \ref{fig:MHDfit}.

\begin{figure*}[ht]
    -\vspace{.6cm}
    \includegraphics[width=.32\textwidth]{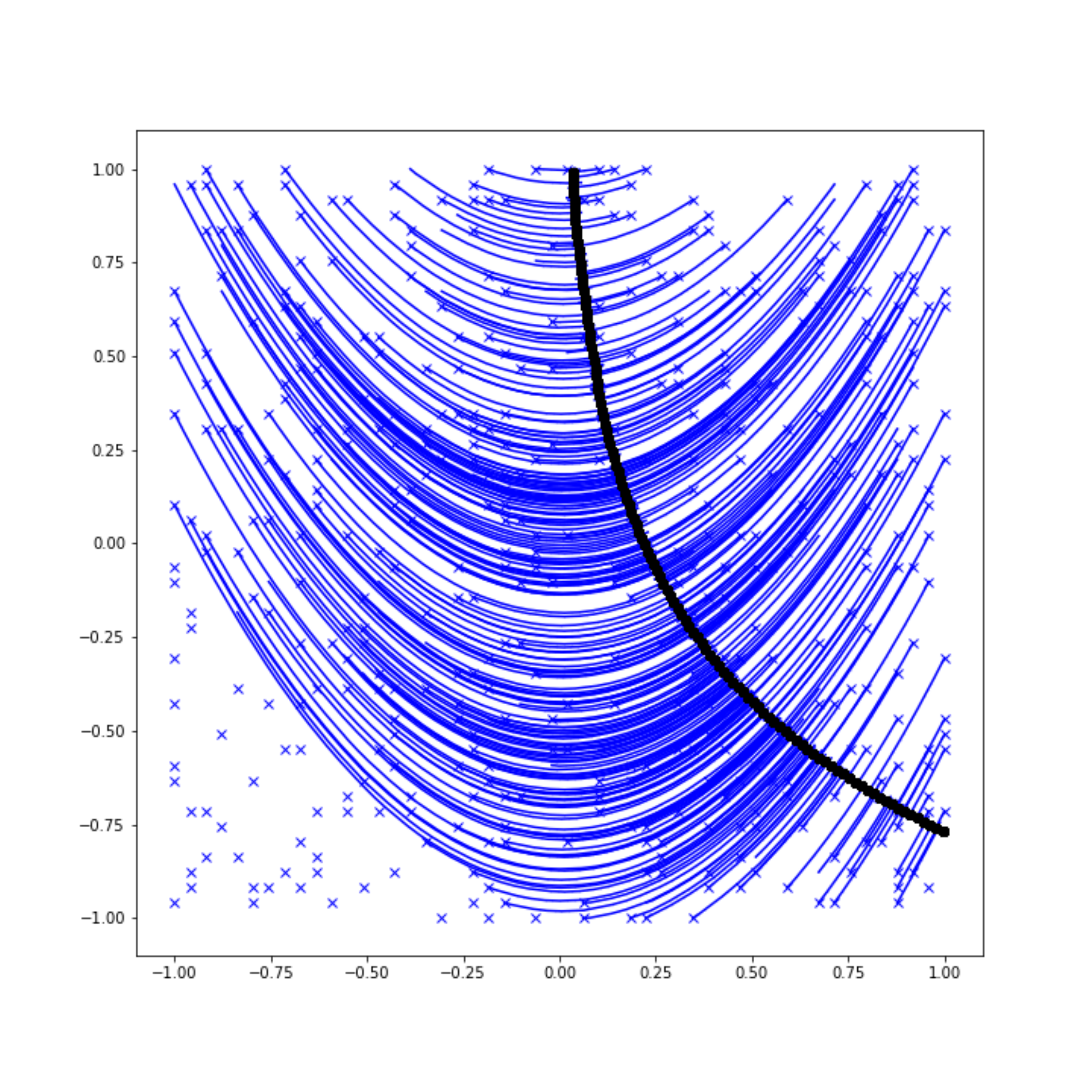}
     \includegraphics[width=.32\textwidth]{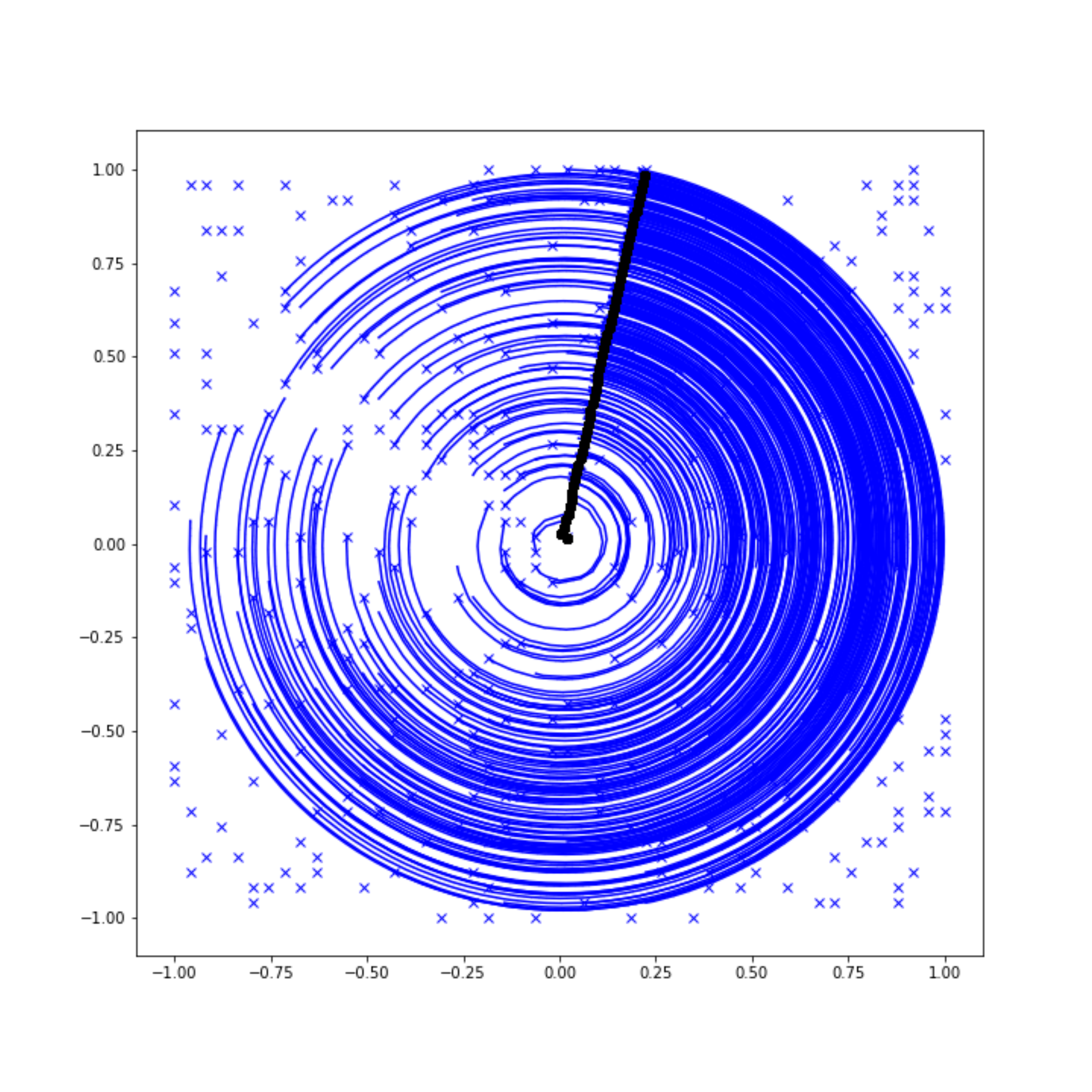}
    \includegraphics[width=.32\textwidth]{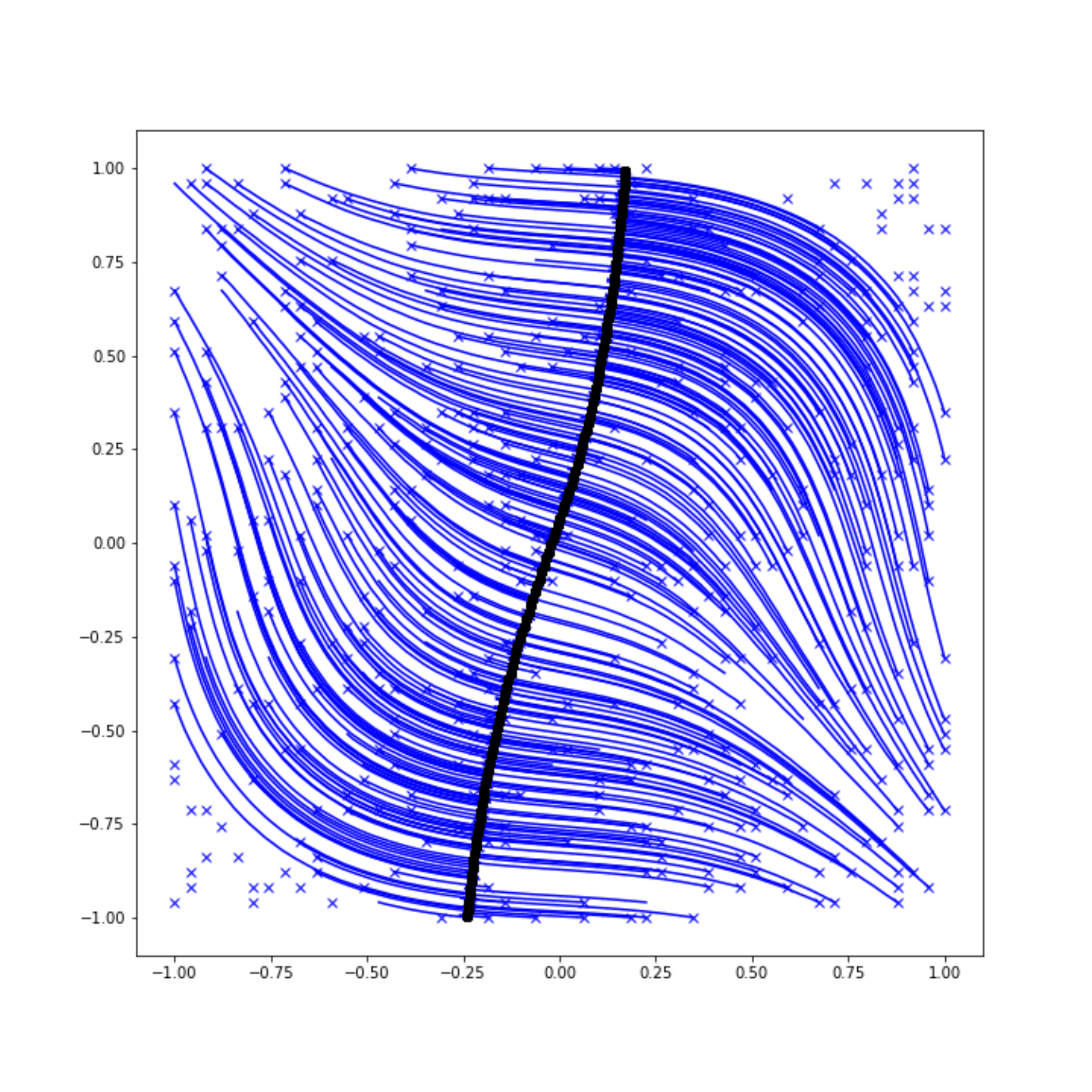}
    \vspace{-2pc}
   \caption{From left to right, AM run on test functions $f_1, f_2, f_3,$ respectively. 
   Black bold line is the approximated active manifold, $\gamma$, from a random starting point as used in tests of Section~\ref{sec:regression-experiment}.
   Test data indicated with \textcolor{blue}{blue $\mathbf{\times}$}, with \textcolor{blue}{path traversed to the active manifold in blue}. Test data with no path indicates those for which the learned active manifold cannot provide an estimate.} 
   \label{fig:am-paths}    
\end{figure*}

\begin{figure}[ht]
  \begin{minipage}[c]{0.32\textwidth}
    \includegraphics[scale=.27]{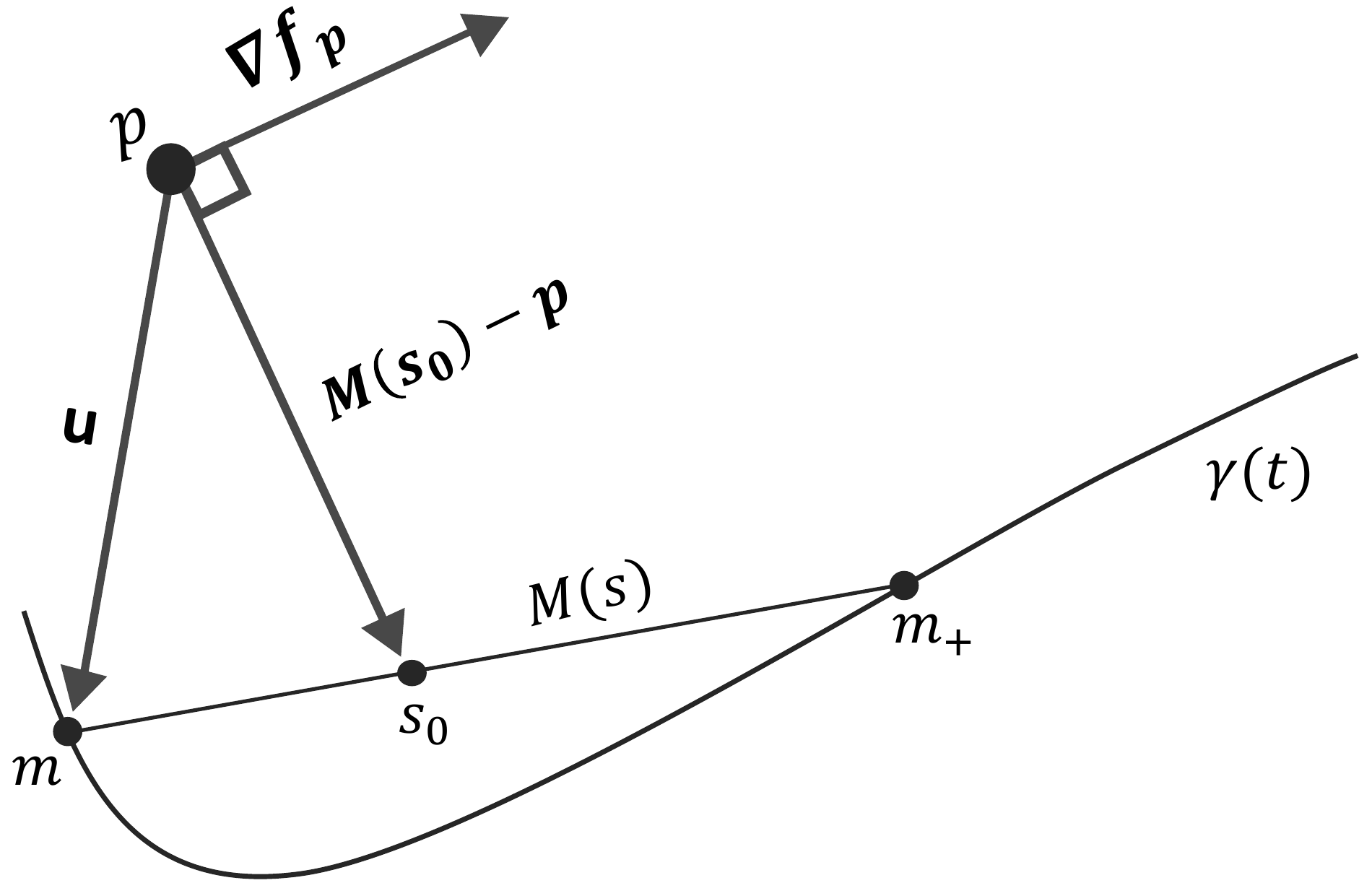}
  \end{minipage}\hfill
  \begin{minipage}[c]{0.16\textwidth}
    \caption{Schematic for the Level Set Algorithm in $\mathbb{R}^2$ when starting point $p$ is one step from active manifold, $\gamma$. Notation in schematic matches pseudo-algorithm.}
 \label{fig:schematic}
  \end{minipage}
\end{figure}
\subsubsection{Traversing the Level Set: }
Given a point $p \in \mathbb{R}^m$, we compute $\hat{f}([p])$ by finding $\gamma(t) \in \mathcal{M}$
 such that $[p] = [\gamma(t)]$. This requires an iterative process that uses vectors orthogonal to $\nabla f_p$ to travel along the level set through $f(p)$ until we intersect the active manifold, $\gamma$. For the following algorithm, we assume that $\nabla f$ has been normalized to unit length, and tolerance $\epsilon$ and step size $\delta$ have been specified.
 
\vspace{-0.6pc}
\begin{enumerate}[leftmargin = *]
\item From $p \in \mathbb{R}^m$, we must identify the direction $\mathbf{v}$ in which to step. First, we try to step greedily toward the active manifold while remaining on the level set. 
Let  $m: = \arg \min |p-m'|$ for $m' \in \mathcal{M}$ (the closest point on the $\mathcal{M}$ to  $p$) and set $\mathbf{u}:= (m-p)/|m-p|$ (the normalized vector from $p$ to $m$). 
We wish to step in the direction of 
\begin{equation}
\label{eq:v}    
\mathbf{v}: = \mathbf{u} - \langle \mathbf{u}, \nabla f_p \rangle \nabla f_p,
\end{equation}
which is the component of $\mathbf{u}$ 
tangent to the level set $f^{-1}(p)$. If $\mathbf{u}$ is not nearly colinear with $\nabla f_p$ (so that $\mathbf{v}$ is not nearly $\vec{0}$), we keep this $\mathbf{v}.$  Usually this is the direction taken for a step. 
On the other hand, if $\mathbf{v}\approx \vec{0}$, walking toward the active manifold would require stepping in line with $\nabla f$, i.e., up/down hill, which we do not permit. 
See Algorithm Notes \ref{sec:notes}.\ref{itm:notes-4} for an if/else loop to define a suitable direction in this case. 

\item Step, $p \mapsto p + \delta \mathbf{v}$. We continue the walk according these first two steps until either (a) we are close to the manifold ($|p-m| \leq \epsilon$), (b) we step out of the hypercube ($p \notin [-1,1]^m$) or (c) the algorithm loops back on itself ($|p-q|\leq \delta/3$ for some previous step $q$). 

\item If $|p-m| \leq \epsilon$, parameterize the line segment, $M(s)$, between $m$ and $m_+$, where $m_+$ is the next closest point on the manifold to $p$. See Figure \ref{fig:schematic} and Alg. Note \ref{sec:notes}.\ref{itm:notes-5}.
    \begin{enumerate}[label*=\arabic*.]
    \item Determine $s_0$ such that $(M(s_0)-p) \perp \nabla f_p$. 

    \item Evaluate $\hat{f}(\gamma(t_0)) = \hat{f}(M(s_0)) \approx f(p)$, where $t_0$ is the corresponding point in $[0,1]$ to $M(s_0)$ via $\gamma$. 
    \end{enumerate}
Else, the traversal along the level set exited the hypercube (or in rare cases self intersected). For these points the chosen active manifold cannot provide an estimate. 
\end{enumerate}


\vspace{-0.5pc}
We note that compactness of $[-1,1]^m$ implies the algorithm necessarily terminates after finite steps based on our stopping conditions. 
For points $p$ near $\mathcal{M}$, continuity of $\nabla f$ implies our algorithm will walk from $p$ approximately along a level set to the necessary intersection with the $\mathcal{M}$. 

\vspace{-0.25pc}
\subsection{Challenges}
Some active manifolds are better choices then others.  
There are regions in the hypercube that will not admit an approximation by the algorithm above, specifically if $p$ is on a level set that is either disconnected or exits the hypercube before reaching an approximated active manifold.  In this case, no approximation is guaranteed; e.g., see Figure~\ref{fig:am-paths}.
In testing, we note that the regions of the hypercube for which our algorithm does not produce an approximation give insight into where it is most useful to build a second active manifold for further approximation.
Hence, iteratively running this algorithm allows for a ``smart''  choice for subsequent active manifold starting points.  Bootstrapping techniques are commonly used for this purpose, placed at the beginning of algorithms to avoid problematic starting configurations. This may provide a worthwhile addition in our case as well.

\subsubsection{Algorithm Notes}
\label{sec:notes}
\begin{enumerate}[leftmargin = *,topsep=5pt,itemsep=2pt,parsep=5pt]
    \item\label{itm:notes-1}
    We have avoided discussing critical points and singularities in the domain of $f$, because we do not yet have effective methods to deal with these obstructions. Currently, it is possible for the algorithm to ``get stuck" near a critical point $c_0$, forcing the user to restart the algorithm on the ``other side" of $c_0$ in order to get a complete active manifold through the domain. With that said, this is not difficult to avoid in practice, and we have written our implementation so that there is no practical risk of the algorithm running indefinitely near these points.

    \item\label{itm:notes-2}
    In practice, one may be given a non-uniform sampling $\{a_i\}_{i=1}^n \subset \mathbb{R}^m$, along with corresponding  values $f(a_i)$ and  $\nabla f_{a_i}$, instead of values on the parameter grid. In this case, the data is scaled linearly to the cube $[-1,1]^m$ and the algorithm is applied as usual.

    \item\label{itm:notes-3}
    One may be enticed to minimize a distance function $D(t) = |\gamma(t)-p|^2$ through differentiation, i.e.
    \vspace{-0.4pc}
    \begin{equation}
        0  = \frac{1}{2}\frac{d}{dt} D(t) = \big\langle \gamma(t)-p, \gamma'(t) \big\rangle
            \vspace{-0.4pc}
    \end{equation}
    But this is not computationally efficient, as we must evaluate both $\gamma'(t)$ and $\gamma(t) - p$. Instead, we recommend computing the values of $|\gamma(t)-p|^2$ and searching for the minimum using another, more efficient algorithm. 
    
    \item \label{itm:notes-4} In the case that $\mathbf{v}\approx \vec{0}$ we proceed in order through the following alternative definitions of $\mathbf{v}$: 
        \vspace{-0.25pc}
        \begin{enumerate}[label*=\arabic*.,topsep=5pt,itemsep=2pt,parsep=5pt]
        \item  If $p-m$ is colinear with $\nabla f_p$, we try to leverage momentum, stepping in roughly the direction of the previous step. 
        Set $\mathbf{u}: = (p_{i}-p_{i-1})/|p_{i}-p_{i-1}|$, and $\mathbf{v}$ as in Eq. \ref{eq:v}. 
        If this is the first step or $\mathbf{v} \approx \vec{0}$ still, proceed to the next bullet.  
        \item Next, we attempt to step towards the origin to prevent walking out of the hypercube. 
        Redefine $\mathbf{u} = p - \vec{0}$ and redefine $v$ from Eqn. \ref{eq:v}. 
        \item  If $v \approx \vec{0}$ still, choose an arbitrary vector in $\nabla f_p^\perp$.
    \end{enumerate}

    \item \label{itm:notes-5} We may express
    $M(s) = (m_+ - m)s + m, \ s \in [0, 1]$ so that the point $s_0$ on $M$ for which $(M(s_0) - p) )\perp \nabla f_p$ can be determined by solving for $s$ in
    \vspace{-0.5pc}
    $$\big\langle (m_+ - m)s + (m-p), \nabla f_p \big\rangle = 0.$$
    \vspace{-0.5pc}
    We then have
    \vspace{-0.5pc}
    $$s_0 = \frac{\big\langle p-m, \nabla f_p \big\rangle}{\big\langle m_+-m, \nabla f_p \big\rangle}\,.$$
    Finally, recall $m, m_+$  are the $k^{th}, (k+1)^{st}$ steps along the $\gamma$, respectively, (Sec. \ref{sec:build-am}), and $m, m_+$ are identified with $k/N, (k+1)/N \in [0,1]$. To apply the spline approximation $\hat f(\gamma(t))$, form the bijective linear map $[k/N, (k+1)/N] \mapsto [m, m_+]$ and choose $t_0 \in [k/N, (k+1)/N]$ corresponding to $s_0 \in [m, m_+]$. 
\end{enumerate}

\section{Examples \& Experiments}
For proof of concept and comparison to AS method in Constantine \citeyear{constantine2015active}, we apply both our AM  and AS to data synthesized from the following two-dimensional test functions  
\begin{align}
\label{eq:example-functions}
    & f_1(x, y) = e^{y - x^2} \\
    & f_2(x,y) = x^2 + y^2 \\
    & f_3(x, y) = x^3 + y^3 + 0.2x + 0.6y.
\end{align}
These functions were chosen to illustrate benefits/problems of the method and to be easy to understand and visualize to facilitate verification and validation. 
We also consider a model magnetohydrodynamic (MHD) power generator considered by Glaws et al. \citeyear{glaws2017dimension} to which they applied AS, to see what extra information we can extract with AM.
{Results code on \url{https://github.com/bridgesra/active-manifold-icml2019-code}. }

\subsection{Two-dimensional Test Functions: }
\label{sec:regression-experiment}
\begin{figure}[ht]
\vspace{-0.5pc}
  \begin{minipage}[c]{0.27\textwidth}
    \includegraphics[width=\textwidth]{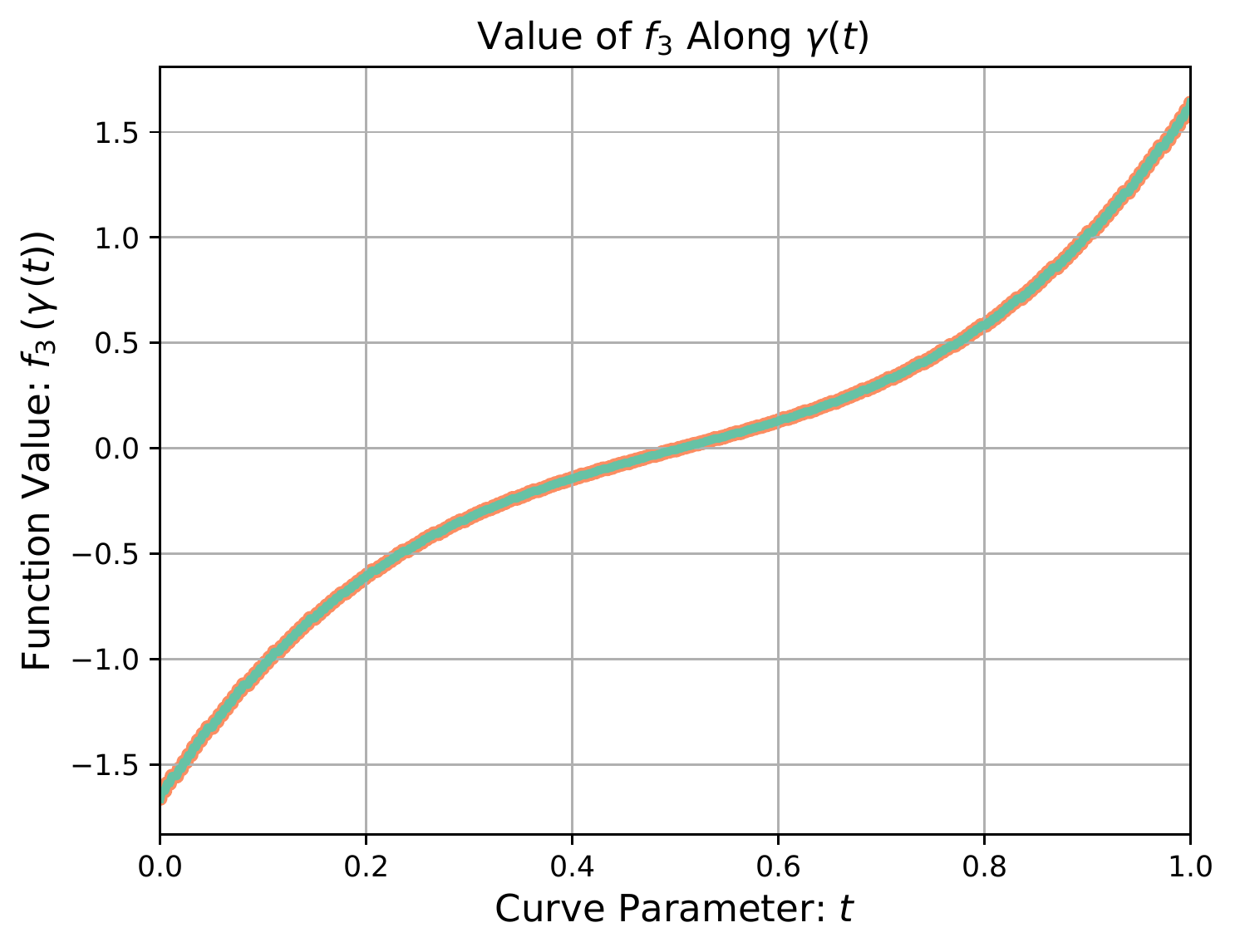}
  \end{minipage}
  \begin{minipage}[c]{0.2\textwidth}
    \includegraphics[width=\textwidth]{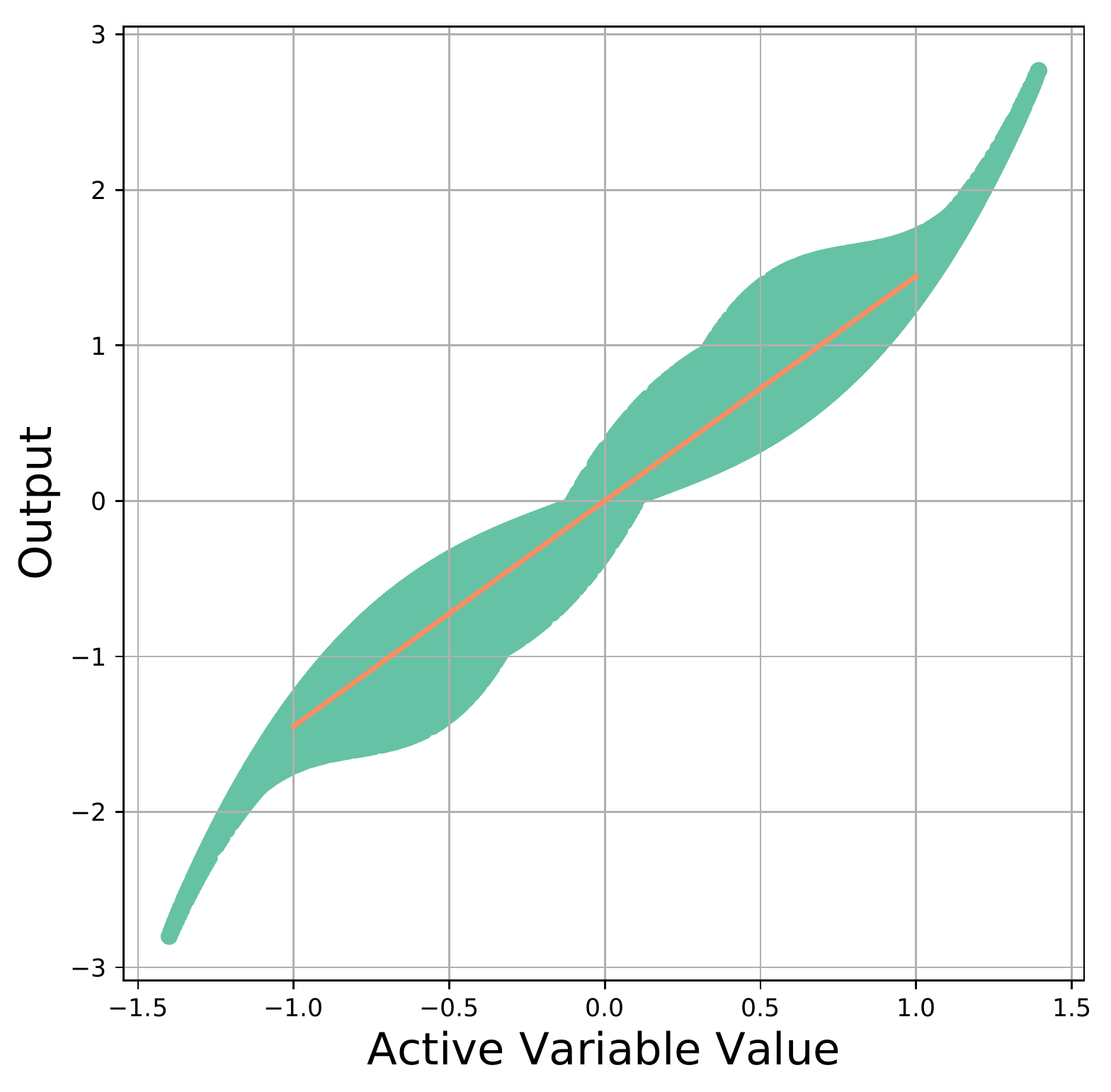}
  \end{minipage}
  \vspace{-0.5pc}
  \caption{(Left) \textit{AM}: Plots of $f_3$ along the active manifold (\textcolor{PineGreen}{green}) and piecewise-cubic Hermite spline approximation from AM  (\textcolor{RedOrange}{orange}). (Right) \textit{AS}: Plots of $f_3$ along the active subspace and bootstrap replicates (\textcolor{PineGreen}{green}), and Constantine's optimization algorithms were applied to fit a degree 4 polynomial (\textcolor{RedOrange}{orange}).}
  \label{fig:f3amas}
\end{figure}

We are interested in how well the AS and AM approximations recover the values of a function for arbitrary points inside the domain. 
For each example function, we approximate with AS and AM and calculate the average error over a set of random test points.
The following steps were followed for each experiment: 
\vspace{-0.25pc}
\begin{enumerate}[leftmargin = *,topsep=3pt,itemsep=2pt,parsep=3pt]
\item A uniform grid $P = \{ p_i \}$ of 10K points was built on $[-1, 1]^2$, with a random 80/20\% partition for training/testing.
\item The values $f(p_i)$ and $\nabla f_{p_i}$ were computed analytically at each grid point, and gradients were normalized so that $|\nabla f_{p_i}|=1$.
\item The active subspace and active manifold were built from the training data $\{p_i,\nabla f_{p_i}\}$, using Constantine's software package \cite{AScode} for the AS, and our algorithm for the AM. (Note that AS necessarily requires the original, unnormalized gradients in this step.)
\item The approximate function $\hat f$ was fit to the resulting data in each case. 
For AM, a piecewise-cubic Hermite interpolation to $f$ was used. 
For AS, Constantine's own optimization algorithm was applied, trained with 100 bootstrap replicates and using a degree-4  polynomial approximation. See Figure~\ref{fig:f3amas}.
\item The 2000 testing points were projected for AS orthogonally to the active subspace, and for AM to the active manifold using our algorithm 3.2.3. The approximation $\hat f$ was then applied to the projected points to generate the values $\hat{f}(p_i)$ for $p_i$ in the training set.
\item The average absolute ($\ell^1$) error and average $\ell^2$ approximation errors in $f$ were computed for each method by comparing the approximate values coming from $\hat f$ against the known analytic values from $f$.
\end{enumerate}

\begin{table}[t]
\resizebox{.49\textwidth}{!}{%
\centering
\begin{tabular}{|c|c|c|c|c|c|c|}
\hline
      &    & \multirow{2}{*}{$\ell^1$ \textbf{mean}} & \multirow{2}{*}{$\ell^1$ \textbf{std}} & \multirow{2}{*}{$\ell^2$ \textbf{mean}} & \multirow{2}{*}{$\ell^2$ \textbf{std}} & $n/N$ \\
      &&&&&& \textbf{mean} \\
\hline
\multirow{2}{*}{$f_1$} & AM &      6.739E-3 &     6.826E-4 &      1.879E-4 &     1.847E-5 &       86.7\% \\ \cline{2-7}
      & AS &         0.585 &     8.130E-3 &         0.751 &     8.600E-3 &        100\% \\
\cline{1-7}
\multirow{2}{*}{$f_2$} & AM &        0.0158 &     9.697E-4 &      4.015E-4 &     2.562E-5 &         77\% \\\cline{2-7}
      & AS &         0.395 &     5.484E-3 &         0.488 &     6.890E-3 &        100\% \\
\cline{1-7}
\multirow{2}{*}{$f_3$} & AM &        0.0106 &     8.442E-4 &      3.154E-4 &     2.887E-5 &       92.9\% \\\cline{2-7}
      & AS &         0.982 &        0.018 &          1.22 &       0.0224 &        100\% \\
\hline
\end{tabular}}
\vspace{-0.8pc}
\caption{Regression results for AM and AS run on test functions $f_1, f_2, f_3$. 
Mean and standard deviation reported across 9 runs (3 train/test splits $\times$ 3 initial AM starting points) for 
average $\ell^1$, $\ell^2$ errors. 
Also we report the ratio $n/N$ of test points  for which the algorithm successfully found an approximation. 
See Figure~\ref{fig:am-paths} for a visualization of one AM run per function.
\vspace{0.25pc}
}
\label{tab:regression}
\end{table}

See Table \ref{tab:regression} for the results. 
Note that AM reduces the average absolute and average $\ell^2$ errors by at least an order of magnitude over AS in each case.
Our initial implementation
is computationally naive and requires more computation than Constantine's AS package \cite{AScode}. In particular, 
AM was on average an order of magnitude slower than AS for these 2-D functions, and this gap increased when testing on a higher dimensional grid of points. 
Further, due to the nonlinear nature of AM, we suspect that AM will never be as fast as AS, though it could be sped up significantly with algorithmic engineering. 
We provide initial performance results of AM and AS in the Supplemental Section. 

\vspace{-.3cm}

\vspace{-0.5pc}
\subsection{MHD Power Generator Model}
We now revisit the work of Glaws et al.~\citeyearpar{glaws2017dimension}, which applied AS to a model for magnetohydrodynamic (MHD) power generation, and consider the application of AM for sensitivity analysis. 

\vspace{-0.25pc}
\subsubsection{The Hartmann Problem}
\label{sec:hartmann}
We first consider the so-called Hartmann problem, which models laminar flow between two parallel plates. 
Following Glaws et al. \citeyearpar{glaws2017dimension}, we examine separately the average flow velocity $u_{avg}$ and the induced magnetic field $B_{ind}$, whose solutions are given analytically in terms of the parameters summarized in Table \ref{MDHHartmannTable}. In particular,
\vspace{-0.2pc}
$$ u_{avg} = -\frac{\partial p_0}{\partial x}\frac{\eta}{B_0^2}\left(1 - \frac{B_0l}{\sqrt{\eta\mu}}\coth\left(\frac{B_0l}{\sqrt{\eta\mu}}\right)\right),$$
\vspace{-0.5pc}
$$B_{ind} = \frac{\partial p_0}{\partial x}\frac{l\mu_0}{2B_0}\left(1 - 2\frac{\sqrt{\eta\mu}}{B_0l}\tanh\left(\frac{B_0l}{2\sqrt{\eta\mu}}\right)\right). $$

\begin{table}[b]
\centering
\resizebox{.48\textwidth}{!}{%
\centering
\begin{tabular}{|l|l|l|l|}
\hline
\multirow{2}{*}{\textbf{Variable}} & \multirow{2}{*}{\textbf{Notation}} & \textbf{Range} & \textbf{Range} \\ 
 & & \textbf{(Hartman)} & \textbf{(MHD)} \\ \hline
Fluid Viscosity & $\log(\mu)$ & $\log([.05, .2])$ & $\log([.001, .01])$\\ \hline
Fluid Density & $\log(\rho)$ & $\log([1, 5])$ & $\log([.1, 10])$\\ \hline
Applied Pressure Gradient & $\log(\frac{\partial p_0}{\partial x})$ & $\log([.5, 3])$ & $\log([.1, .5])$\\ \hline
Resistivity & $\log(\eta)$ & $\log([.5, 3])$ & $\log([.1, 10])$\\ \hline
Applied Magnetic Field & $\log(B_0)$ & $\log([.1, 1])$ & $\log([.1, 1])$ \\ \hline
Magnetic Constant & $\mu_0$ & fixed at 1 & fixed at 1 \\
\hline
Length & $l$ & fixed at 1  & fixed at 1\\
\hline
\end{tabular}%
}
\vspace{-1pc}
\caption{Parameters and ranges for the Hartmann and idealized MHD problem reproduced from Glaws et al. ~\citeyearpar{glaws2017dimension}. The first five are variable, while the others are fixed at 1.}
\label{MDHHartmannTable}
\end{table}

\vspace{-0.25pc}
To analyze this probem using AM, the following experiment was conducted analogous to Glaws et al.:
\begin{enumerate}[leftmargin = *,topsep=4pt,itemsep=2pt,parsep=4pt]
    \item The cube $[-1,1]^5$ was discretized with a uniform grid of $14^5$ evenly-spaced points, and these points were mapped linearly through a dilation map $D$ onto their appropriate ranges as given in Table \ref{MDHHartmannTable}.
    \item Function values $u_{avg},\, B_{ind}$, and their gradients were computed analytically on these inputs from the formulae provided by Constantine \citeyearpar{ASdata}.
    \item The computed gradients were mapped back to the cube $[-1,1]^5$ through $D^{-1}$ (taking the chain rule into account) and normalized to unit length.
    \item The AM algorithm was run on this data from random seed 46 with $\delta = \epsilon = $ 0.02.
\end{enumerate}

\begin{figure}[t]
    \centering
    \begin{subfigure}[b]{.49\textwidth}
        \centering
        \includegraphics[width=.475\linewidth]{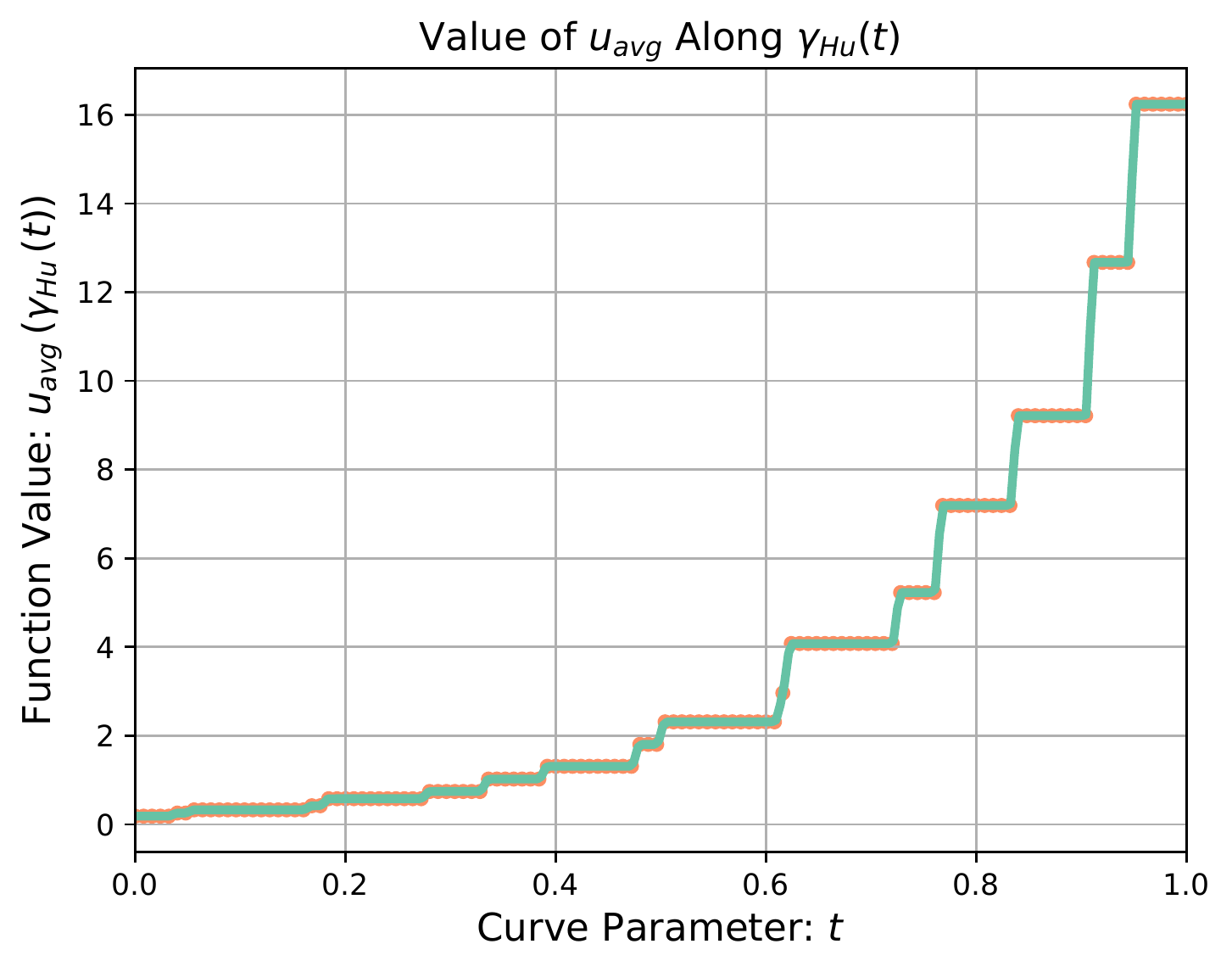}
        \hfill
        \includegraphics[width=.475\linewidth]{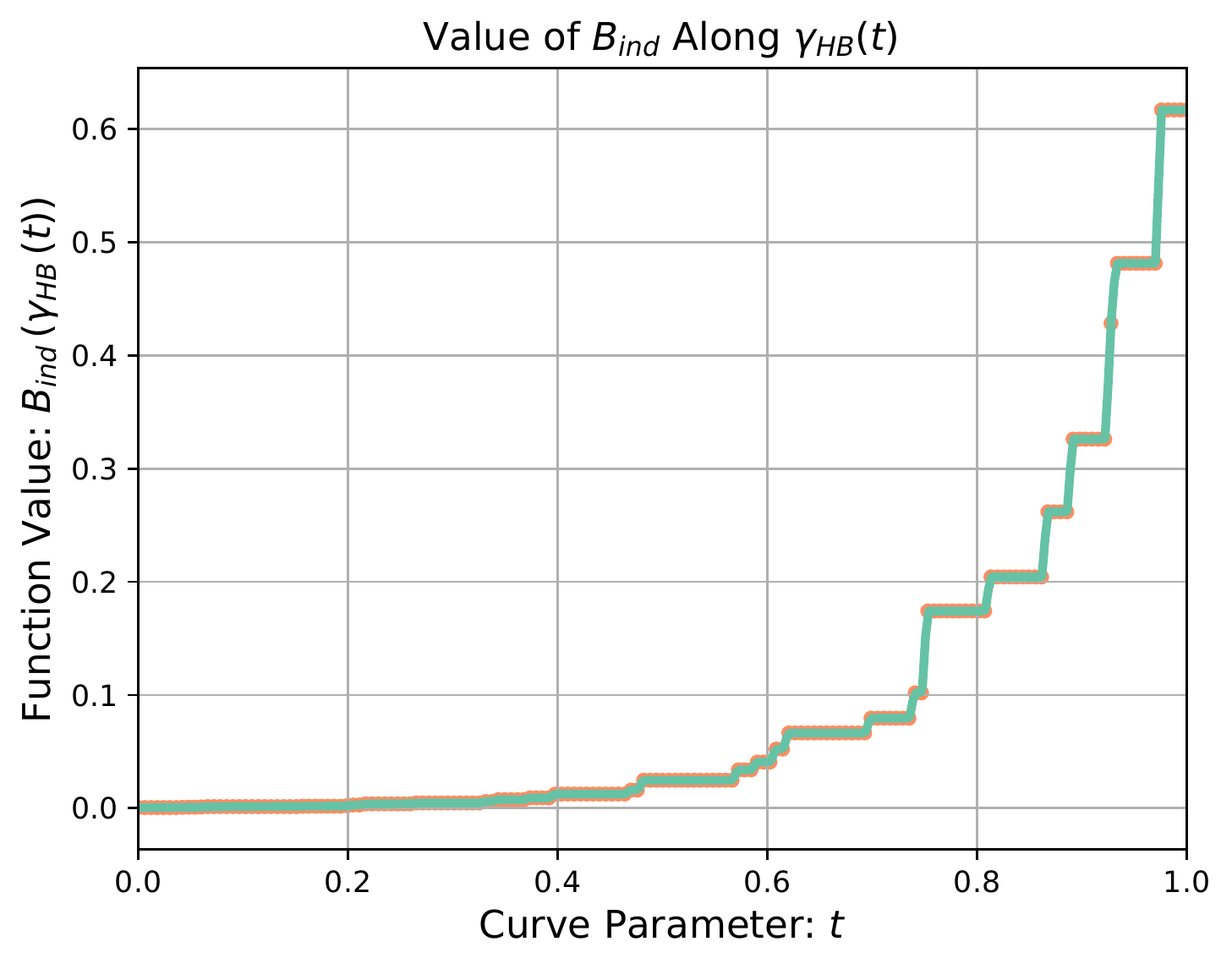}
    \end{subfigure}
    \begin{subfigure}[b]{.49\textwidth}
        \centering
        \includegraphics[width=.475\linewidth]{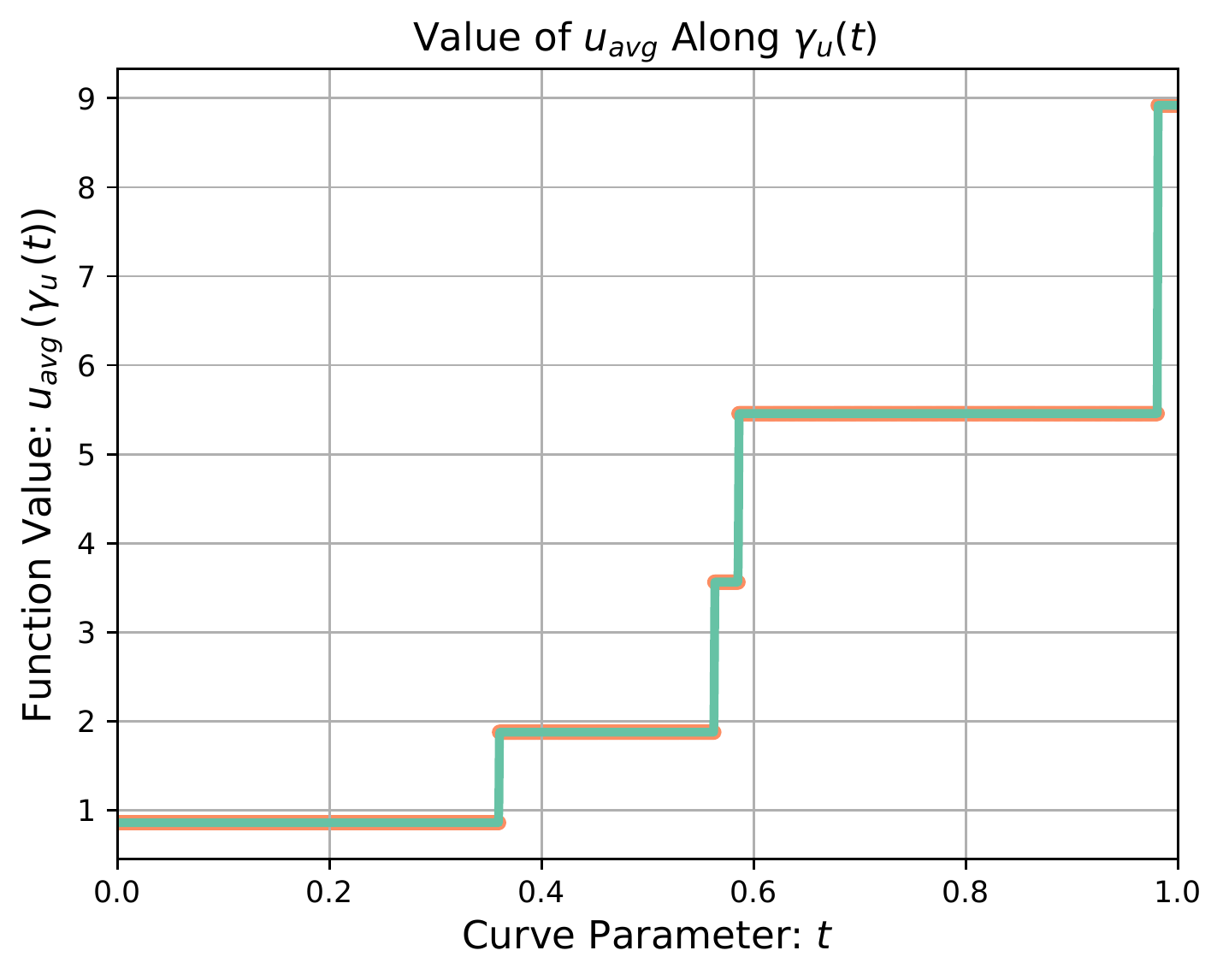}
        \hfill
        \includegraphics[width=.475\linewidth]{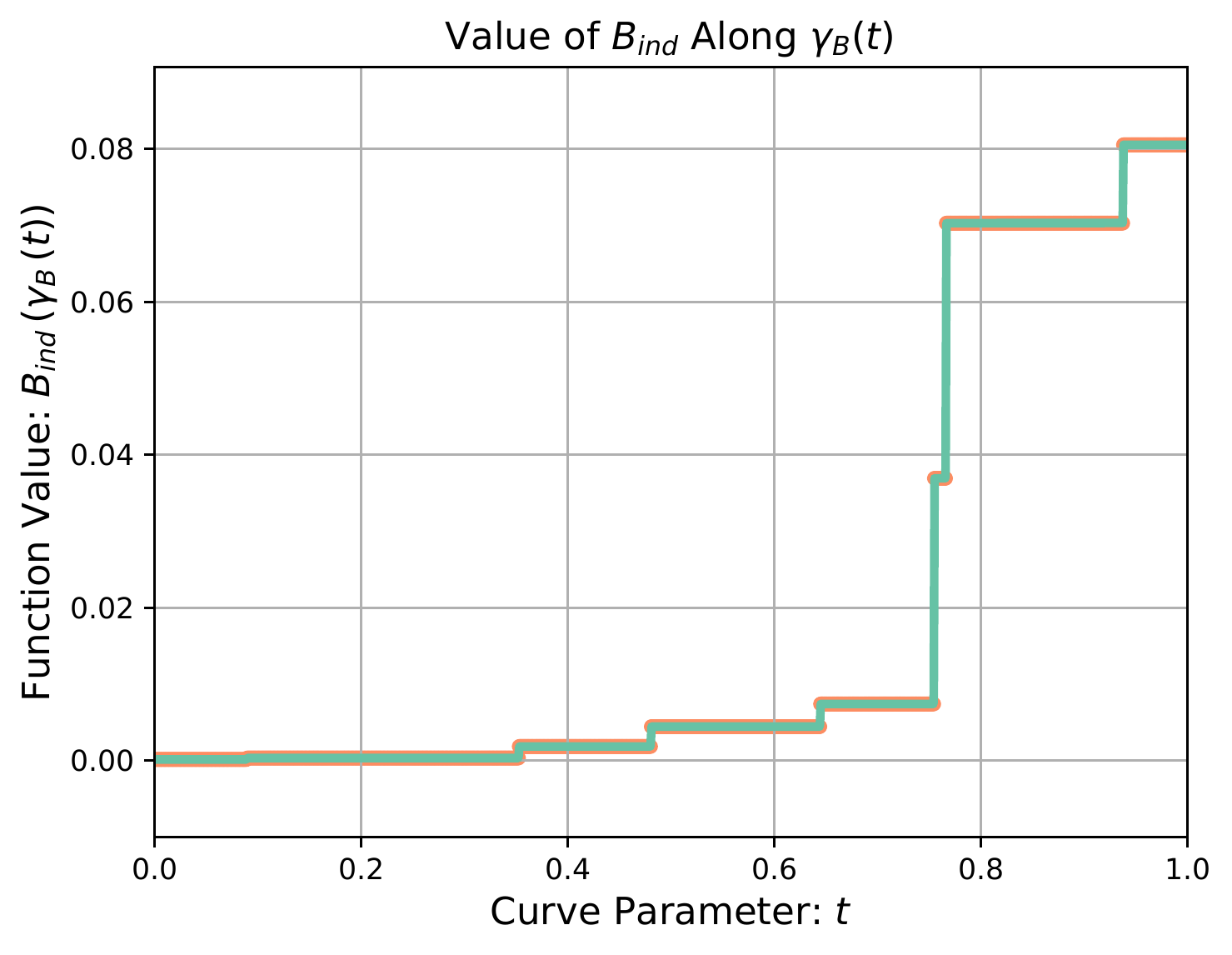}
    \end{subfigure}
    \vspace{-1.5pc}
    \caption{
    AM-derived function values in \textcolor{RedOrange}{orange} along parameterization of the active manifold  and piecewise-cubic Hermite interpolation fit to these points in \textcolor{PineGreen}{green}.   
    Top plots depict $u_{ave}$ (left), $B_{ind}$ (right) corresponding to the Hartmann problem (Sec. \ref{sec:hartmann}).  Bottom plots depict  $u_{ave}$ (left), $B_{ind}$ (right) corresponding to the idealized MHD generator data (Sec. \ref{sec:ideal-mhd}).}  
    \label{fig:MHDfit}
    \vspace{-.4cm}
\end{figure}
\begin{figure}[!ht]
    \vspace{-2mm}
    \centering
    \begin{subfigure}[b]{.49\textwidth}
        \centering
        \includegraphics[width=.475\linewidth]{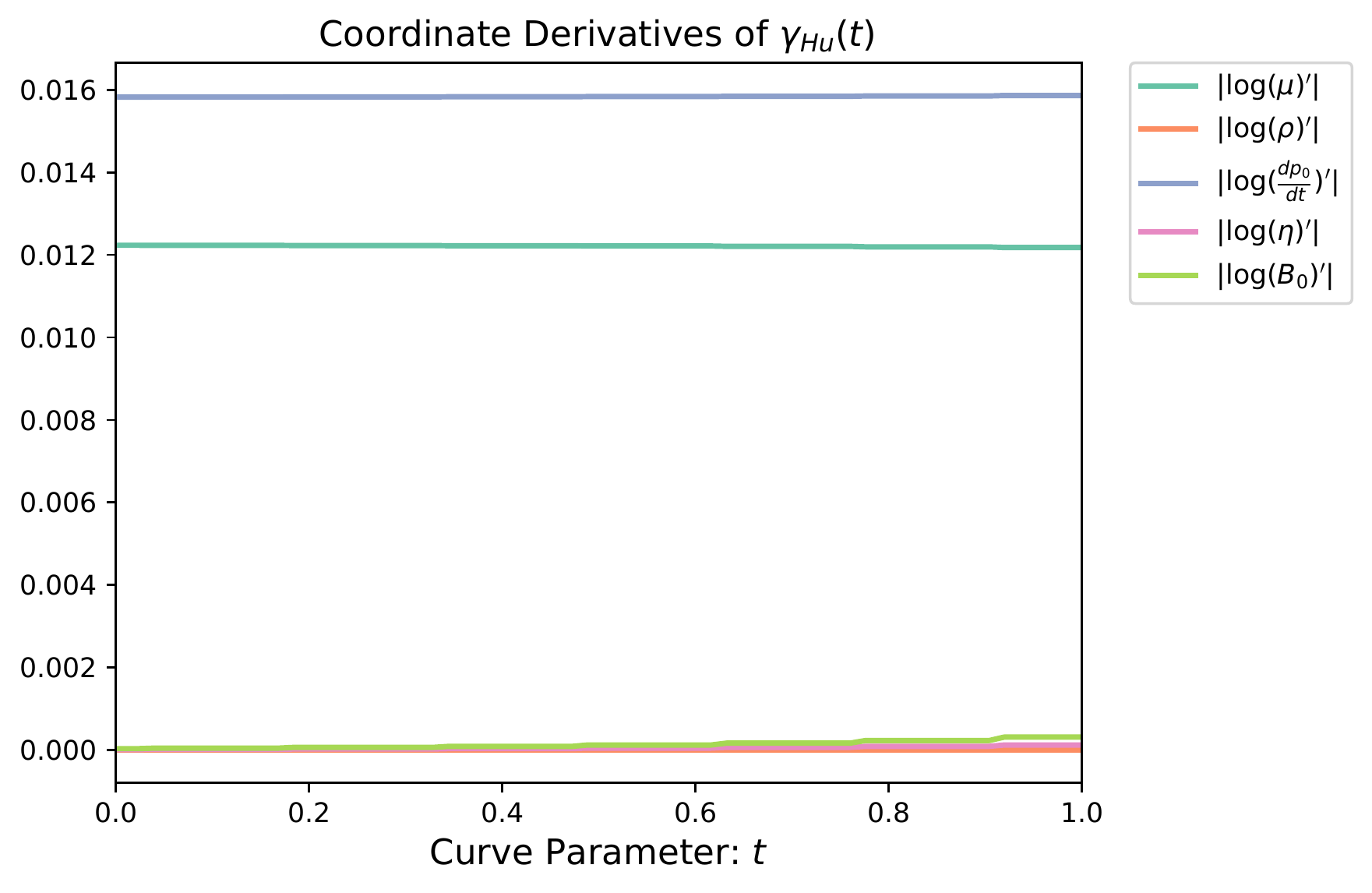}
        \hfill
        \includegraphics[width=.475\linewidth]{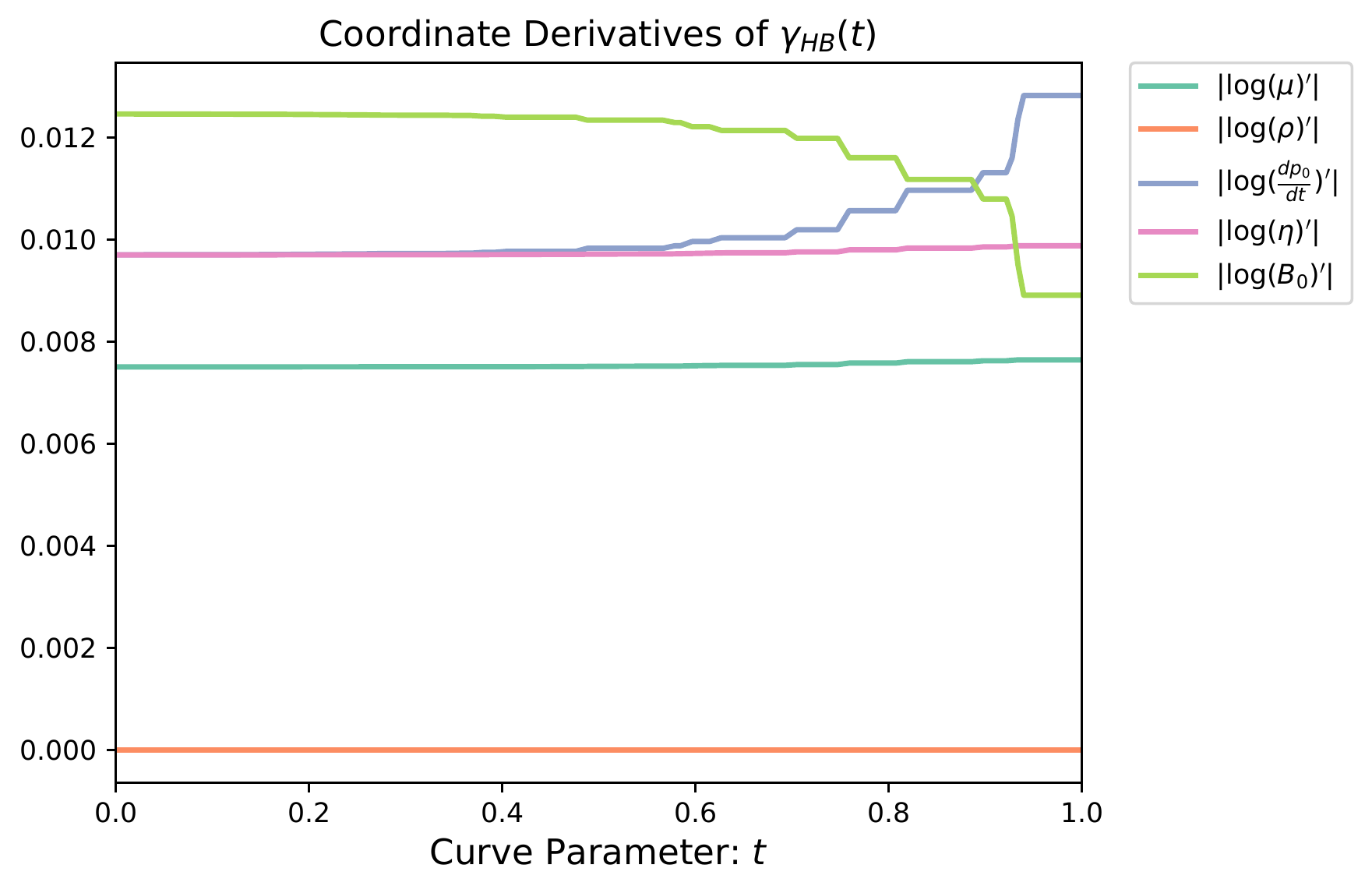}
    \end{subfigure}
    \begin{subfigure}[b]{.49\textwidth}
        \centering
        \includegraphics[width=.475\linewidth]{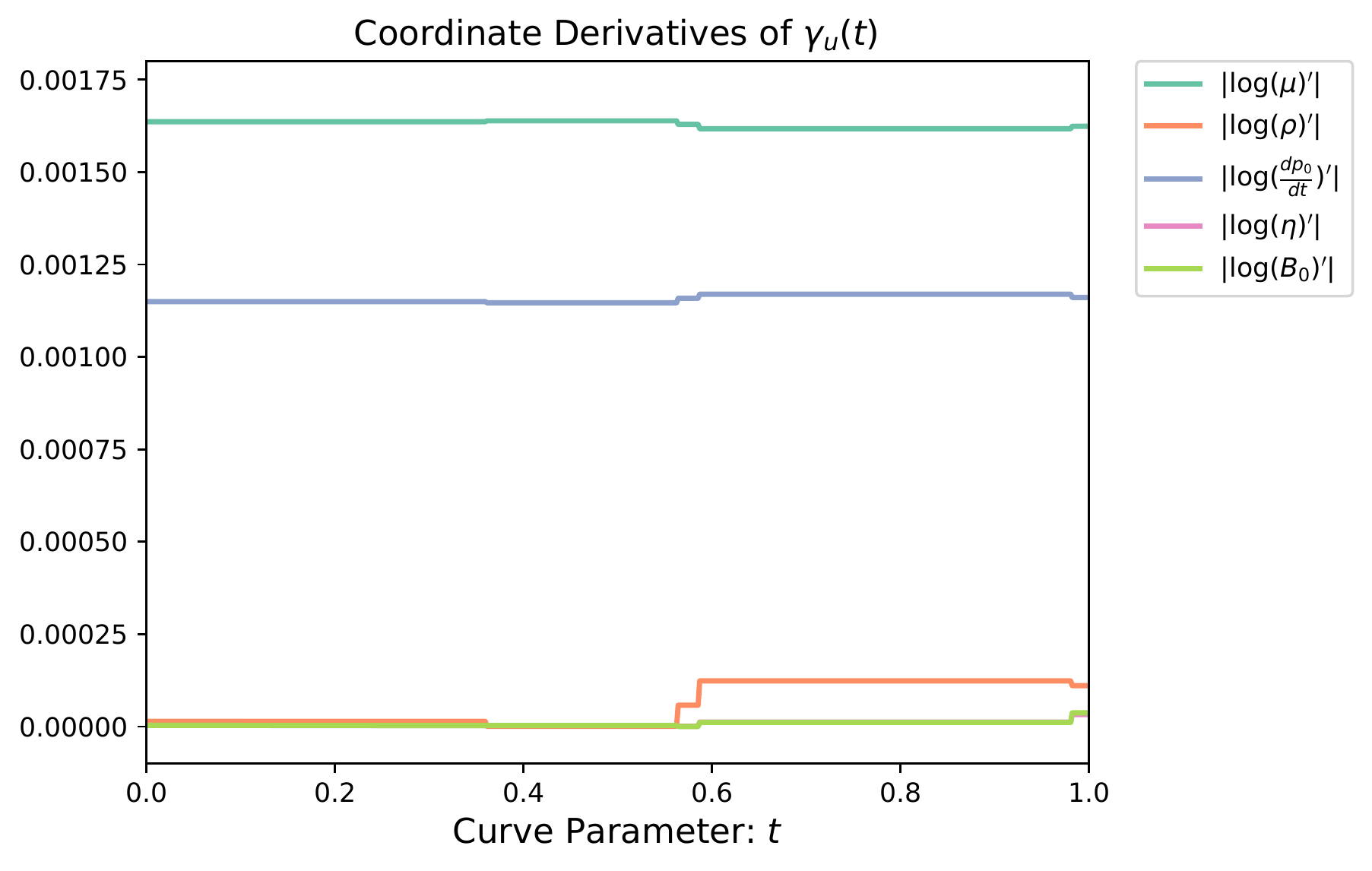}
        \hfill
        \includegraphics[width=.475\linewidth]{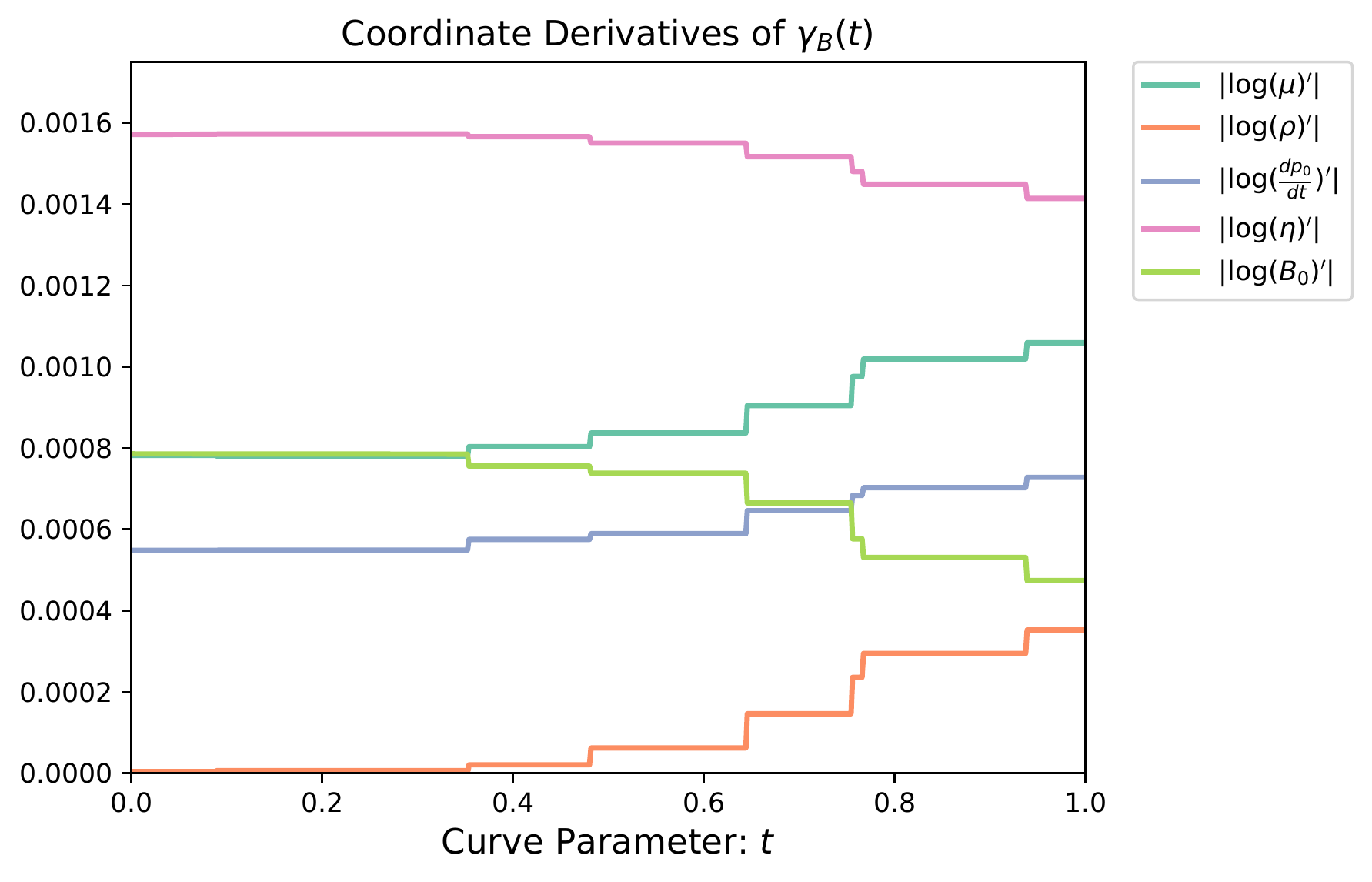}
    \end{subfigure}
    \caption{Plots of the magnitude of each parameter (partial) derivative  along the active manifolds $\gamma_{(\cdot)}$. Top:  $\gamma_{Hu}$ (left) and $\gamma_{HB}$ (right) corresponding to the Hartmann problem (Sec. \ref{sec:hartmann}). Bottom: $\gamma_{u}$ (left) and $\gamma_{B}$ (right) corresponding to the idealized MHD model (Sec. \ref{sec:ideal-mhd}). Note that the AM permits analysis of sensitivity of input variables and especially their changes along the active manifold, which is not possible with AS. }
    \label{fig:MHDderivs}
\end{figure}

The fit is nearly exponential in both cases\textemdash see the top row of Figure~\ref{fig:MHDfit}\textemdash  which is expected given that $\coth{x}$ and $\tanh{x}$ are rational functions of an exponential variable. 
We also begin to see why one may prefer our method to AS in some situations, as the top row of Figure~\ref{fig:MHDderivs} shows that there is nonlinear behavior in the derivatives along the active manifold $\gamma_{HB}$ corresponding to  $B_{ind}$ that will be missed with an affine model like AS.
Note that Glaws et al. remark in \citeyearpar{glaws2017dimension} that for both quantities of interest, a 2-D affine subspace is sufficient to almost completely characterize the output, which is believable given our results. 

However, we also see from Figure~\ref{fig:MHDderivs} that 
the relative influence of the parameters on $B_{ind}$ changes for parameter configurations near the last quarter of the active manifold. 
Indeed, the applied pressure gradient (\textcolor{blue}{blue}) begins to overtake both the resistivity (\textcolor{Rhodamine}{pink}) and the previously-dominant applied magnetic field (\textcolor{LimeGreen}{lime}). This behavior is reasonable, as the $\tanh(x)$ function multiplying $\eta$ ``levels off" as $|x|$ increases, so the $\partial p_0 / \partial x$ term in the equation for $B_{ind}$ begins to take precedence.

For further analysis, we also compared AS to AM on this data\textemdash following the same procedure as for the 2-D test functions. 
We constructed a uniform grid of 100K points on $[-1,1]^m$, and ran AM using stepsize 0.15 over three random 98K\,/\,2K test/train splits, each with three initial AM start points. We then ran AS on the same data using Constantine's software package \cite{AScode}, along with his optimization algorithms as before. Results are displayed in Table \ref{tab:HartmannASvAM}, again showing increased accuracy of AM over AS. 


\begin{table}[t]
\vspace{-.05cm}
\centering
\begin{tabular}{| c | c | c | c | c |}
\hline
& \multicolumn{2}{c|}{$B_{ind}$} & \multicolumn{2}{c|}{$u_{avg}$} \\
\cline{2-5}
&        AM &        AS &        AM &     AS \\
\hline
$\ell^1$ mean &    0.0367 &     0.154 &      1.09 &   4.87 \\
\hline
$\ell^1$ std  &  3.063E-3 &  5.883E-3 &     0.255 &  0.103 \\
\hline
$\ell^2$ mean &  1.286E-3 &     0.244 &     0.033 &   7.02 \\
\hline
$\ell^2$ std  &  3.415E-4 &    0.0116 &  3.774E-3 &  0.163 \\
\hline
\end{tabular}
\caption{Hartmann data average approximation errors reported. 
As expected, we see a large improvement with AM for $B_{ind}$ where there is nonlinear behavior and less improvement for $u_{avg}$ where parameter influence remains static.}
\label{tab:HartmannASvAM}
\end{table}

\subsubsection{Idealized MHD Generator}
\label{sec:ideal-mhd}
The situation becomes even more interesting when we apply AM to the set of data provided by Glaws et al.~\citeyearpar{glaws2017dimension}.  
This data is used to study a model for idealized 3D duct flow through a MHD generator. 
It involves the same parameters considered in the Hartmann problem, but with different ranges, as found in Table \ref{MDHHartmannTable}. 
The procedure for processing the 483 samples of inputs and outputs was similar to that for the Hartmann problem.  
The AM algorithm was applied with random seed 46 and step sizes $\delta = \epsilon = $ 0.002.

Examining the plot in Figure~\ref{fig:MHDderivs} of the derivatives along the active manifold, $\gamma_B$, we see a large amount of change in the influence of each parameter throughout the parameter space.  
Fluid viscosity (\textcolor{teal}{blue-green}) and applied magnetic field (\textcolor{LimeGreen}{lime})  have the same amount of effect on the output $B_{ind}$ at the beginning of $\gamma_B$, but from the middle to end of $\gamma_B$,  fluid viscosity increases influence while applied magnetic field decreases. 
Further, applied pressure gradient (\textcolor{blue}{blue}) overtakes applied magnetic field. 
Overall, this shows the influence of each parameter and how it varies along the active manifold. 
This represents an important case of behavior that is undetectable by AS.
Since AS produces global sensitivity rankings as linear combinations of parameters, AS fails to capture the behavior of each parameter individually or changes in influence through the space. 


\section{Conclusions}
We provide mathematical background and introduce a novel algorithm, AM, for analyzing $C^1$ functions $f$ with a problematic ratio of inputs to observations. 
Leveraging initial samples of $f$ and $\nabla f$, our algorithm computes an approximate gradient streamline and recovers the function on this 1-D submanifold.  
Specifically, from a given point of interest $p$, the algorithm traverses the level set of $f$ through $p$, intersecting the active manifold at a point that allows for accurate approximation. 
We provide initial tests on known functions for both accuracy and performance, showing the algorithm outperforms AS in accuracy at greater computational expense. 
Further, we demonstrate the efficacy of AM in parameter studies using the data and MHD generator model considered by Glaws et al. \citeyearpar{glaws2017dimension} to demonstrate the AS method. 
Our results permit deeper understanding of the effect of each parameter on the function, specifically showing which parameters are most/least influential along an active manifold, and where this sensitivity changes.

\section*{Acknowledgements}
Special thanks to Guannan Zhang and the reviewers whose comments helped polish this paper. This work was supported in part by the U.S. Department of Energy, Office of Science, Office of Workforce Development for Teachers and Scientists (WDTS) under the program SULI and the National Science Foundation's Math Science Graduate Research Internship.

\small

\bibliography{am.bib}
\bibliographystyle{icml2019}

\section{Supplemental Section: Performance of Initial Implementation}
\label{sec:supplement}

The following experiment was run to provide a timing comparison, using $f(x) = |x|^2$ as the test function and a 2013 Macbook Pro with 16GB of RAM and a 2.4 GHz Intel I7:
\begin{enumerate}[leftmargin = *,topsep=4pt,itemsep=2pt,parsep=4pt]
    \item A uniform grid of dimension $m$ was constructed, consisting of $n$ points in each dimension. Function values and gradients were then sampled on this grid, with the gradients normalized to unit length. For each of the 3 tests, the data set was randomly partitioned into 3 training/testing sets according to the test proportion in Table \ref{tab:time}, and the AM was built on the training set using 3 random initial points.
    \item Step size for AM was chosen to be $2/3$ times the length of the longest grid diagonal i.e. $2/3*(1/n)\sqrt{m}$. Execution time was recorded.
    \item AS was run on the data (with un-normalized gradients) and execution time was recorded.
\end{enumerate}
We note as an aside that error estimates in both AS and AM remained relatively unchanged despite variation in these experimental parameters. The execution time comparison is shown in Table \ref{tab:time}.

\begin{table}[!h]
\centering
\begin{tabular}{|l|l|l|l|l|}
\hline
$m$                & $n$                 & \begin{tabular}[c]{@{}l@{}}Test\\ Fraction\end{tabular} & AM time & AS time \\ \hline
\multirow{4}{*}{2} & \multirow{2}{*}{15} & 1/6                                                     & 324ms   & 21.9ms  \\ \cline{3-5} 
                  &                     & 1/3                                                     & 522ms   & 20.0ms  \\ \cline{2-5} 
                  & \multirow{2}{*}{30} & 1/6                                                     & 2.62s   & 24.7ms  \\ \cline{3-5} 
                  &                     & 1/3                                                     & 5.61s   & 25.1ms   \\ \hline
\multirow{4}{*}{3} & \multirow{2}{*}{15} & 1/6                                                     & 5.17s     & 50.6ms   \\ \cline{3-5} 
                  &                     & 1/3                                                     & 10.9s   & 60ms   \\ \cline{2-5} 
                  & \multirow{2}{*}{30} & 1/6                                                     & 120s         & 606ms       \\ \cline{3-5} 
                  &                     & 1/3                                                     & 246s       & 1.64s       \\ \hline
\end{tabular}
\caption{\small{Some simple execution time results for AS vs AM. Main takeaway is AM is consistently an order of magnitude slower than AS. }}
\label{tab:time}
\end{table}



\end{document}